\documentclass[10pt,twocolumn,letterpaper]{article}

\usepackage[pagenumbers]{cvpr} 

\usepackage{graphicx}
\usepackage{amsmath}
\usepackage{amssymb}
\usepackage{booktabs}

\usepackage{multicol}
\usepackage{multirow}
\usepackage{xcolor}
\usepackage{colortbl}

\definecolor{cvprblue}{rgb}{0.21,0.49,0.74}
\definecolor{secondgreen}{HTML}{36B449}
\usepackage[pagebackref,breaklinks,colorlinks,citecolor=cvprblue]{hyperref}

\def\paperID{4275} 
\def\confName{CVPR}
\def\confYear{2024}

\title{Video-Bench: A Comprehensive Benchmark and Toolkit for Evaluating Video-based Large Language Models}

\author{
    Munan Ning$^{1,2}$\thanks{Equal contribution.}, Bin Zhu$^{1}$\footnotemark[1], Yujia Xie$^{3}$, Bin Lin$^{1}$,  Jiaxi Cui$^{1,4}$, Lu Yuan$^{3}$, Dongdong Chen$^{3}$\thanks{Corresponding author}, Li Yuan$^{1,2}$\footnotemark[2] \\ 
    $^1$ Peking University \quad
    $^2$ PengCheng Laboratory \quad
    $^3$ Microsoft \quad
    $^4$ FarReel AI Lab
    }

\begin{document}
\maketitle

\begin{abstract}
Video-based large language models (Video-LLMs) have been recently introduced, targeting both fundamental improvements in perception and comprehension, and a diverse range of user inquiries.
In pursuit of the ultimate goal of achieving artificial general intelligence, a truly intelligent Video-LLM model should not only see and understand the surroundings, but also possess human-level commonsense, and make well-informed decisions for the users. 
To guide the development of such a model, the establishment of a robust and comprehensive evaluation system becomes crucial. To this end, this paper proposes \textit{Video-Bench}, a new comprehensive benchmark along with a toolkit specifically designed for evaluating Video-LLMs. The benchmark comprises 10 meticulously crafted tasks, evaluating the capabilities of Video-LLMs across three distinct levels: Video-exclusive Understanding, Prior Knowledge-based Question-Answering, and Comprehension and Decision-making. In addition, we introduce an automatic toolkit tailored to process model outputs for various tasks, facilitating the calculation of metrics and generating convenient final scores.
We evaluate 8 representative Video-LLMs using \textit{Video-Bench}. 
The findings reveal that current Video-LLMs still fall considerably short of achieving human-like comprehension and analysis of real-world videos, offering valuable insights for future research directions.
The benchmark and toolkit are available at: \url{https://github.com/PKU-YuanGroup/Video-Bench}.
\end{abstract}

\section{Introduction}
\label{sec:intro}
\begin{figure}[t]
    \centering
    \includegraphics[width=1\columnwidth]{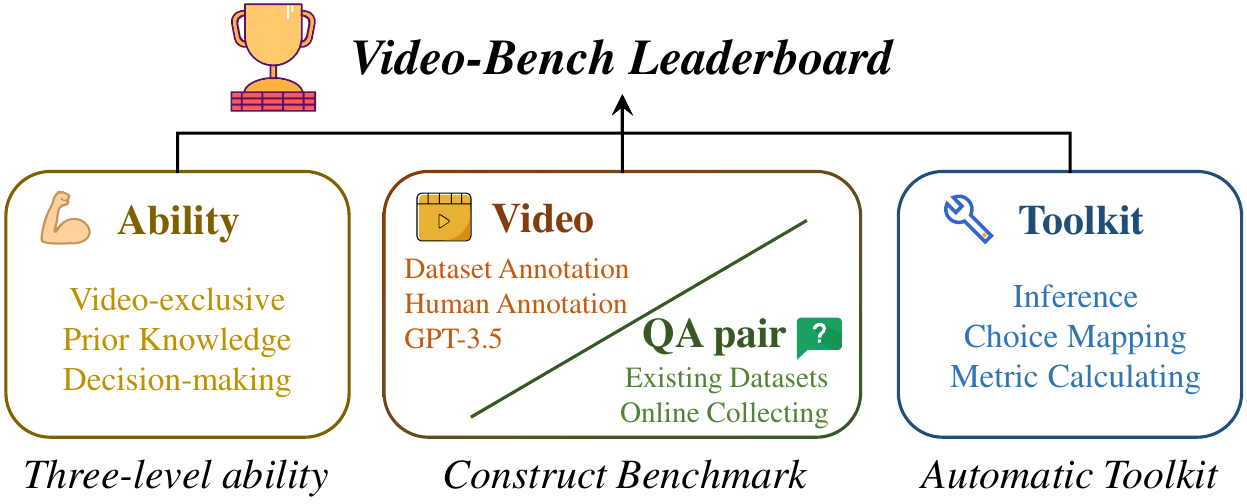}
    \caption{The illustrative pipeline for the intuition, construction and application of \textit{Video-Bench}.}
    \label{fig:title}
    \vspace{-1.5em}
\end{figure}

\begin{table*}[htpb]
    \caption{Comparison between different Video-LLMs. `VE', `TM', `AE', `LLM', and `Adapt' denote the visual encoder, temporal module, audio encoder, LLM backend and the adaptation module. The `CLIP (L)' and `CLIP (O)' represent the CLIP encoder pre-trained on LLaVA~\cite{liu2023visual} and OpenFlamingo~\cite{awadalla2023openflamingo}. If the models are trained with two-stage, the training data of each stage is split by `/'. The `combined' denotes the combination of typical V-L datasets including COCO~\cite{chen2015microsoft}, CC~\cite{sharma2018conceptual}, VG~\cite{krishna2017visual}, SBU~\cite{ordonez2011im2text} and LAION~\cite{schuhmann2021laion}.}
    \label{table:vllm}
    \centering
    \setlength\tabcolsep{1.8mm}
    \renewcommand\arraystretch{1.0}
    {
    \small 
    \scalebox{0.95}{
    \begin{tabular}{l|ccccc|cc}
        \toprule[1pt]
        \multirow{2}{*}{\textbf{Method}} &\multicolumn{5}{c|}{\textbf{Model Configuration}}  &\multicolumn{2}{c}{\textbf{Training Data}}  \\
         &VE &TM &AE &LLM &Adapt  &Source &Size\\
        \midrule [1pt]
        VideoChat~\cite{li2023videochat} &BLIP-2 &GMHRA &Whisper &Vicuna &Q-Former &Combined / Instruct-video &35M / 18K\\        
        \midrule
        Video-ChatGPT~\cite{maaz2023video} &CLIP (L) &AVG Pool &- &Vicuna &Linear &Instruct-video &100K\\
        \midrule
        Otter~\cite{li2023otter} &CLIP (O) &- &- &LLaMA (O) &Linear &MIMIC-IT &2.8M\\
        \midrule
        PandaGPT~\cite{su2023pandagpt} &ImageBind &- &ImageBind &Vicuna &Linear &LLAVA-mniGPT4 &153.5K\\
        \midrule
        Valley~\cite{luo2023valley} &CLIP (L) &AVG Pool &- &Vicuna &Linear &WebVid / Instruct-video &702K / 47.8K\\
        \midrule
        mPLUG-Owl~\cite{ye2023mplug} &CLIP &- &- &LLaMA &Abstractor &Combined / LLaVA &1100M / 150K\\
        \midrule
        Video-LLaMA~\cite{zhang2023video} &BLIP-2 &Frame Emb &ImageBind &Vicuna &Q-Former &WebVid / LLAVA-mniGPT4 &2M / 153.5K\\
        \midrule
        Chat-UniVi~\cite{Chat-UniVi} & CLIP (L) & Cluster & - & Vicuna & Linear & Combined / Instruct-video & 1.5M / 649K\\
        
        \bottomrule[1pt]
        \end{tabular}}
        }
\end{table*}

Large language models (LLMs)\cite{radford2018improving,radford2019language,brown2020language,ouyang2022training,touvron2023llama,touvron2023llama2} have demonstrated strong capabilities in handling natural language processing (NLP) tasks, including comprehension, composition and reasoning, and achieved remarkable advancements on NLP benchmarks\cite{clark2018think,zellers2019hellaswag,hendrycks2020measuring,lin2021truthfulqa}.
This success has also inspired studies on Video-LLMs~\cite{wang2022omnivl,maaz2023video,li2023videochat,li2023otter,su2023pandagpt,luo2023valley,chen2023x,lyu2023macaw,wang2023chatvideo}, where models process video inputs with textual prompts and generate corresponding answers, shedding light on the future format of artificial general intelligence (AGI) for video understanding.


With the ultimate goal of achieving artificial general intelligence in mind, we assert that a truly intelligent video-language model should at least exhibit three distinct human-like capabilities: (\textit{i}) Video-exclusive Understanding,  i.e., performing well for questions whose answer can be extracted from the video itself; (\textit{ii}) Prior Knowledge-based Question-Answering, i.e., answer questions that require the prior knowledge beyond the video, such as commentary on NBA games or providing background information on specific music videos;  (\textit{iii}) Comprehension and Decision-making, enabling a comprehensive understanding of scenarios, along with the ability to make predictions and informed decisions. Example applications encompass 3D scene understanding and decision-making for autonomous driving.

To gradually approach this goal, the establishment of an evaluation benchmark is indispensable for precisely measuring and steering the development progress.
However, we find that existing benchmarks fall short of serving this purpose comprehensively.
For instance, MMBench~\cite{liu2023mmbench} and LVLM-eHub~\cite{xu2023lvlm} are concentrated on image understanding, ignoring the video understanding ability.
SEED-Bench~\cite{li2023seed} includes several video tasks but is limited to temporal understanding, thus only covering the first level. To this end, we propose a new large-scale benchmark along with a toolkit,  referred to as ``\textit{Video-Bench}", to furnish a thorough evaluation of Video-LLMs. The composition of \textit{Video-Bench} is depicted in Fig.~\ref{fig:title}.

In detail, aligning with our motivation, our \textit{Video-Bench} encompasses tasks categorized into three distinct levels of capability: (\textit{i}) For Video-exclusive Understanding, we begin by randomly selecting parts of traditional QA pairs~\cite{xu2017video,yu2019activitynet,jang2017tgif}, and proposing more challenging tasks to assess both temporal and contextual aspects of videos. Tasks include video summarization~\cite{zhou2018towards}, abnormal detection~\cite{sultani2018real}, and crowd counting~\cite{leal2015motchallenge};
(\textit{ii}) For Prior Knowledge-based Question-Answering, we evaluate the capability of model in understanding TV dramas~\cite{lei2018tvqa}, appreciating music videos, and providing information about players and games in NBA videos.
(\textit{iii}) For Comprehension and Decision-making, we employ two classical tasks: 3D indoor scene understanding~\cite{ma2022sqa3d} and auto-driving decision-making to assess the comprehension and decision-making abilities of models.

To streamline the evaluation process, we include another crucial component, i.e., the evaluation toolkit, along with the benchmarks. The toolkit automatically maps the long text outputs of Video-LLMs to corresponding answers with probability selection~\cite{hendrycks2020measuring} or LLM-based semantic understanding~\cite{ouyang2022training,raffel2020exploring}. Subsequently, it calculates accuracy for each question and generates a final score, enhancing the efficiency of the evaluation workflow.

We evaluate eight representative Video-LLMs on \textit{Video-Bench}: VideoChat~\cite{li2023videochat}, Video-ChatGPT~\cite{maaz2023video},
Otter~\cite{li2023otter}, Valley~\cite{luo2023valley}, PandaGPT~\cite{su2023pandagpt}, mPLUG-Owl~\cite{ye2023mplug}, Video-LLaMA~\cite{zhang2023video}, and Chat-UniVi~\cite{Chat-UniVi} with verified open-source model weights. The evaluation results reveal several interesting findings: (\textit{i}) Most recent models can summarize the main content of videos but lack the capacity to detect details and temporal information. (\textit{ii}) Due to the absence of domain-specific prior knowledge in the training data, these models encounter challenges in accurately comprehending and responding to queries within a particular domain. 
(\textit{iii}) 
Due to constraints in multimodal information extraction and the use of a weakened LLM backend (either 7B or 13B), the majority of tested models exhibit limited proficiency in comprehending and decision-making within complex scenarios. Our contributions can be summarized as follows: 

\begin{itemize}
\item We introduce \textit{Video-Bench}, the first comprehensive evaluation benchmark for Video-LLMs, featuring a three-level ability assessment that systematically evaluates models in video-exclusive understanding, prior knowledge incorporation, and video-based decision-making abilities.
\item We provide a user-friendly evaluation toolkit. Accompanied by our datasets and QA pairs, the toolkit can streamline the performance assessment of Video-LLMs.
\item We conduct extensive experiments to evaluate prominent Video-LLMs, summarizing their behaviors, analyzing main causes for observed limitations, and proposing future directions for improvement.
\end{itemize}

\begin{figure*}[ht]
    \centering
    \includegraphics[width=2\columnwidth]{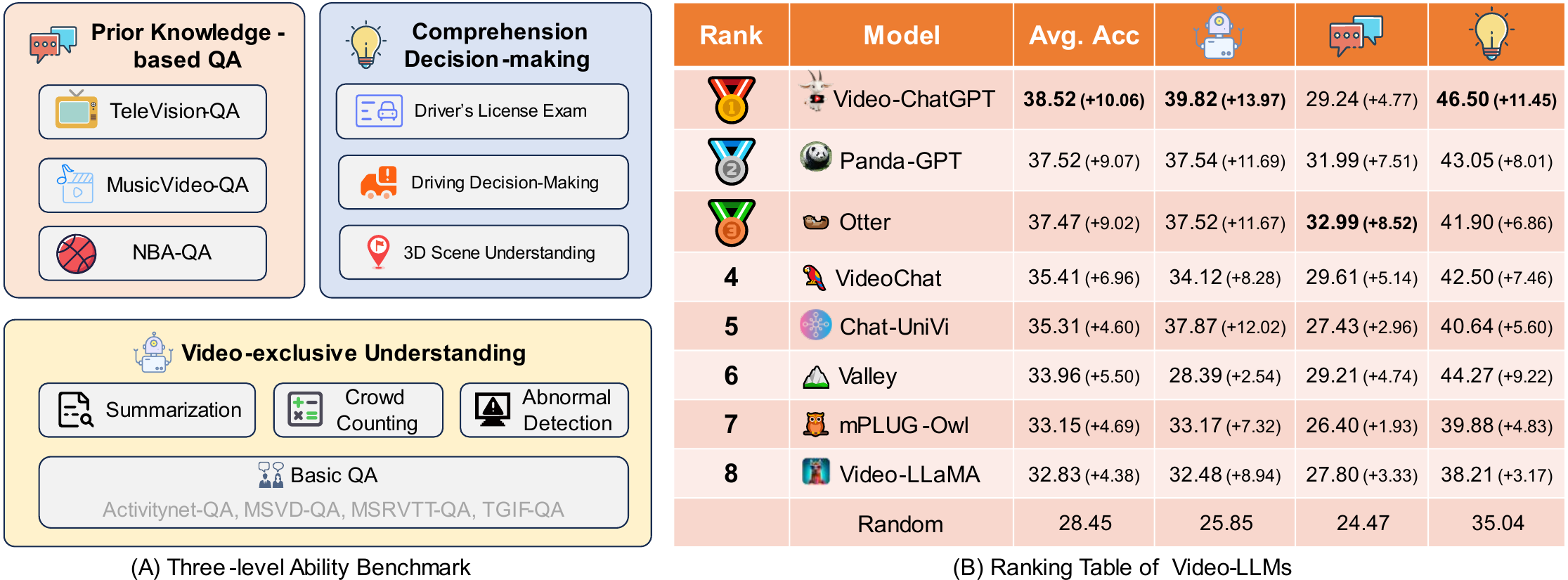}
    \caption{(A) part denotes the three-level evaluation of our \textit{Video-Bench}. (B) part denotes the ranking of existing Video-LLMs. The reported results are accuracy (\%) calculated with GPT-3.5. The number in parentheses represents the improvement over random results.}
    \label{fig:main}
\end{figure*}

\section{Related Work}
\noindent\textbf{Video-LLMs.}
Extending Image-based Large Language Models (Image-LLMs) to the video modality introduces a complex challenge, necessitating the incorporation of temporal dimensions to interpret diverse frame information. Beyond visual content, the integration of audio, subtitles, and other modalities becomes crucial for a comprehensive understanding of video semantics. In response to this challenge, a series Video-LLMs have emerged, building upon open-source LLMs ~\cite{touvron2023llama,touvron2023llama2,chiang2023vicuna} or Image-LLMs~\cite{alayrac2022flamingo,awadalla2023openflamingo,liu2023visual,yuan2021florence}.

As outlined in Table~\ref{table:vllm}, VideoChat~\cite{li2023videochat} utilizes the Q-Former to map visual representations to Vicuna~\cite{chiang2023vicuna}, implementing a two-stage training process. Video-ChatGPT~\cite{maaz2023video} and Valley~\cite{luo2023valley} originate from the LLaVA~\cite{liu2023visual} framework and introduce average pooling to enhance temporal sequence perception. Otter~\cite{li2023otter} proposes the MIMIC-IT dataset and fine-tunes Openflamingo~\cite{awadalla2023openflamingo} on their dataset. PandaGPT~\cite{su2023pandagpt} employs the ImageBind~\cite{girdhar2023imagebind} as its backend for video comprehension. mPLUG-Owl~\cite{ye2023mplug} introduces an abstractor module to align image and text. Video-LLaMA~\cite{zhang2023video} incorporates a frame embedding layer and ImageBind to inject temporal and audio information into the LLM backend, while Chat-UniVi~\cite{Chat-UniVi} merges visual tokens with similar semantic meanings using a clustering strategy.
Existing Video-LLMs vary in their training strategies and data scales, with only a subset addressing challenges related to temporal dimensions and audio modalities.

\vspace{0.1cm}
\noindent\textbf{Video Datasets.}
Deep learning for video analysis relies on diverse datasets tailored to specific tasks. A notable task is human action recognition, featuring action classification datasets such as UCF-101~\cite{soomro2012ucf101}, HMDB51~\cite{kuehne2011hmdb}, and Kinetics~\cite{kay2017kinetics}, and action localization datasets like AVA~\cite{gu2018ava} and Fineaction~\cite{liu2022fineaction}. Tasks involving anomaly detection in surveillance videos are addressed by datasets like UCSD-anomaly~\cite{mahadevan2010anomaly} and UCF-crime~\cite{sultani2018real}. Object identification and tracking in videos encompass multiple object tracking (MOT)\cite{leal2015motchallenge}, video object segmentation (DAVIS)\cite{perazzi2016benchmark}, and video instance segmentation (Youtube-VIS)~\cite{yang2019video}.
For multimodal tasks, video captioning datasets such as MSVD~\cite{chen2011collecting}, MSRVTT~\cite{xu2016msr}, and Activitynet~\cite{caba2015activitynet} exist, along with their corresponding QA datasets~\cite{xu2017video,xu2017video,yu2019activitynet}. Scenario-specific datasets like MovieQA~\cite{tapaswi2016movieqa} and TVQA~\cite{lei2018tvqa} also contribute to the diversity of available datasets. However, these datasets often focus on specific tasks and lack the complexity needed to measure the comprehensive abilities of Video-LLMs effectively.

\vspace{0.1cm}
\noindent\textbf{Vision Language Evaluation Benchmarks.}
To evaluate the capabilities of LLMs, various benchmarks have been introduced, including AI2 Reasoning~\cite{clark2018think}, HellaSwag~\cite{zellers2019hellaswag}, MMLU~\cite{hendrycks2020measuring}, and TruthfulQA~\cite{lin2021truthfulqa}. These benchmarks assess reasoning, scientific knowledge, fact retention, and the ability to generate misinformation. In the realm of multimodal LLMs, corresponding benchmarks have also emerged.
MMBench~\cite{liu2023mmbench} constructs a broad spectrum of evaluation for Vision-LLMs, and converts free-form predictions into pre-defined choices, enhancing the robustness of the evaluation process.
SEED-Bench~\cite{li2023seed} introduces a series of temporal understanding tasks and establishes an automatic filtering and manual verification pipeline to ensure the quality and relevance of the evaluations.
LVLM-eHub~\cite{xu2023lvlm} presents an online arena platform for user-level evaluation, providing a more realistic assessment of model performance in real-world applications.
ELEVATER~\cite{li2022elevater} focuses on evaluating the transferability of language-augmented visual models across multiple tasks.
However, the aforementioned vision-language benchmarks are not tailored specifically for videos. Drawing inspiration from HELM~\cite{liang2022holistic}, we introduce \textit{Video-Bench}, specifically designed to measure human-like abilities of Video-LLMs across various capabilities and scenarios.

\begin{figure*}[ht]
    \centering
    \includegraphics[width=2\columnwidth]{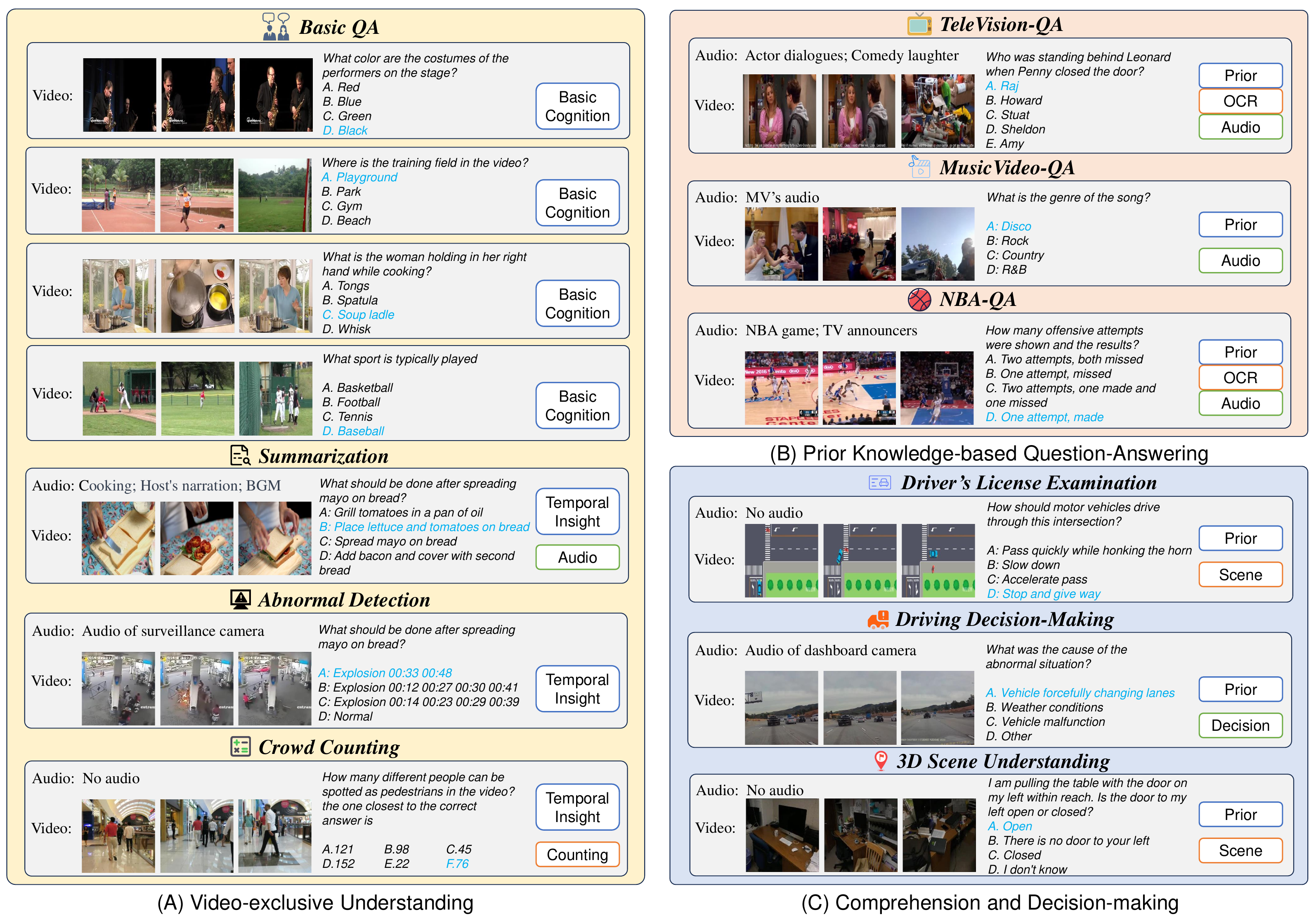}
    \caption{The detailed illustration of samples from each task and the corresponding ability required.}
    \label{fig:visual_ability}
\end{figure*}

\section{Video-Bench}
In Fig.\ref{fig:main}, we show the overall structure of \textit{Video-Bench}  and the corresponding average results for existing Video-LLMs.

\subsection{Video-exclusive Summarization}
\label{sec:video-exclusive}
As illustrated in Fig.~\ref{fig:visual_ability} (A), we aim to measure the capacity of Video-LLMs to comprehend and summarize information from video itself, encompassing objects, actions, attributes, and their temporal connections. These tasks are video-exclusive, requiring no external prior knowledge or complex logic inference.

\vspace{0.1cm}
\noindent\textbf{Basic Understanding.} This task primarily evaluates the basic video recognition ability, such as responding to queries related to human actions in Activitynet-QA~\cite{yu2019activitynet}, providing answers related to objects, attributes, and actions corresponding to videos in MSVD-QA~\cite{xu2017video} and MSRVTT-QA~\cite{xu2017video}, and comprehending GIFs in TGIF-QA~\cite{jang2017tgif}.

\vspace{0.1cm}
\noindent\textbf{Summarization.} This task assesses the summarization ability of Video-LLMs when dealing with longer videos. Using the YouCook2 dataset~\cite{zhou2018towards} with rich annotations and extended video duration, we generate a series of QA pairs to evaluate whether the model can comprehend cooking information presented in the videos and audios, and then provide accurate feedback about the correct procedure.

\vspace{0.1cm}
\noindent\textbf{Abnormal Detection.} This task evaluates the ability to review videos and identify anomalies. Leveraging the UCF-Crime dataset~\cite{sultani2018real}, a collection of surveillance videos annotated with the type and timestamp of anomalies, we construct questions to assess the temporal comprehensive ability of Video-LLMs.

\vspace{0.1cm}
\noindent\textbf{Crowd Counting.} This task primarily evaluates the ability to localize and count dense objects. Utilizing the MOT dataset~\cite{leal2015motchallenge}, which annotates all pedestrians, vehicles, and other targets in street or mall images, we test whether Video-LLMs can identify different pedestrians in different frames and provide the correct number of people.

\subsection{Prior Knowledge-based Question-answering}
\label{sec:prior}
ChatGPT and LLaMA exhibit strong capability in answering questions and giving suggestions across various domains due to the extensive prior knowledge acquired during pre-training. This prompts us to investigate whether Video-LLMs possess similar abilities. As depicted in Fig.~\ref{fig:visual_ability} (B), our goal is to assess the capability of Video-LLMs in addressing questions that require prior knowledge, akin to human beings. Examples include identifying actors in a movie or discerning the music style of a particular song.

\vspace{0.1cm}
\noindent\textbf{TV-QA.} Television programs, as prevalent sources of entertainment videos, integrate multiple modalities, including video, audio, and subtitles, to convey information. Utilizing the TVQA dataset~\cite{lei2018tvqa}, we transform image formats into videos, and incorporate audio and subtitles. This dataset allows us to evaluate the ability of Video-LLMs to integrate prior knowledge and information from video, audio, and text to answer questions related to TV content.

\vspace{0.1cm}
\noindent\textbf{MV-QA.} Music videos, characterized by the synchronization of visual elements with music, pose a unique challenge due to their reliance on prior knowledge. Answering questions about these videos requires familiarity with the song, recognition of artists, and potentially basic music theory. In the absence of relevant existing datasets, we search for top music videos on YouTube and construct corresponding QA pairs based on authoritative wiki sources. This task assesses the ability of Video-LLMs to understand the song associated with the music video and provide answers regarding performers, background information, and relevant music theory knowledge.

\vspace{0.1cm}
\noindent\textbf{NBA-QA.} Understanding competitive sports videos also demands relevant prior knowledge. Viewers must possess knowledge of the corresponding rules and engage in long-term observation to identify competing teams, players, technical actions, scores, or fouls within the video. We select top NBA plays from YouTube and manually annotate teams, players, and technical actions in each game, transforming them into question-answer pairs. These videos and questions serve as input to the model, expecting it to respond based on relevant prior knowledge.

\begin{table*}[htpb]
    \caption{Experiment results of tested Video-LLMs on various tasks. `*' denotes the QA-pairs are re-constructed or annotated by \textit{Video-Bench}. `$^{\dagger}$' denotes the tasks with fewer videos and multiplying the weight by 0.5 when calculating the final result. For each task, \textcolor{blue}{\textbf{blue}} and \textcolor{secondgreen}{green} mark the \textcolor{blue}{\textbf{first}} and \textcolor{secondgreen}{second} place respectively. All the reported results are accuracy (\%) calculated with GPT-3.5-based~\cite{ouyang2022training} metric. The ``Video-'' and ``Chat-'' are abbreviated to ``V-'' and ``C-''.}
    \label{table:main_result}
    \centering
    \setlength\tabcolsep{1.2mm}
    \renewcommand\arraystretch{1}
    {
    \small 
    \scalebox{1.0}{
    \begin{tabular}{l|l|ccccccccc}
        \toprule[1pt]
        \multicolumn{11}{c}{\textbf{(A) Video-Exclusive Understanding}} \\
        \midrule
        Task & Dataset &Random &V-Chat &V-ChatGPT &Otter &PandaGPT &Valley &mPLUG &V-LLaMA &C-UniVi \\
        \midrule
        \multirow{4}{*}{Basic QA} &Activitynet-QA &34.58 &44.55  &\textcolor{secondgreen}{46.60} &44.30 &44.96 & 38.10 &41.47 &39.85 &\textcolor{blue}{\textbf{49.00}} \\
                                  &MSVD-QA* &26.22 &42.15 &\textcolor{blue}{\textbf{57.50}} &\textcolor{secondgreen}{54.95} &50.43 &32.03 &42.45 &41.15 &48.60 \\
                                  &MSRVTT-QA* &26.50 &37.40 &\textcolor{secondgreen}{46.30} &\textcolor{blue}{\textbf{46.95}} &44.60 &28.03 &36.30 &34.05 &41.65 \\
                                  &TGIF-QA &22.37 &33.74 &\textcolor{secondgreen}{35.59} &{34.27} &29.66 &31.41 &31.66 &31.28 &\textcolor{blue}{\textbf{41.30}} \\
        \midrule
        Summarization & YouCook2* &25.00 &27.66 &\textcolor{blue}{\textbf{34.80}}  &32.65 &\textcolor{secondgreen}{33.02} &29.05 &27.05 &28.90 &29.00 \\
        \midrule
        Abnormal Detection & UCF-Cirme* &25.00 &22.41 &24.13 &22.41  &\textcolor{blue}{\textbf{33.01}} &20.34 &22.76 &{27.59} &\textcolor{secondgreen}{28.27} \\
        \midrule
        Crowd Counting & MOT*$^{\dagger}$ &16.67 &\textcolor{blue}{\textbf{27.78}} &\textcolor{blue}{\textbf{27.78}} &\textcolor{secondgreen}{16.67} &\textcolor{secondgreen}{16.67} &11.11 &\textcolor{blue}{\textbf{27.78}}  &\textcolor{secondgreen}{16.67} &\textcolor{secondgreen}{16.67}\\
        \midrule
        \multicolumn{2}{c|}{Average Score} &25.85 &34.12 &\textcolor{blue}{\textbf{39.82}} &37.52 &37.54 &28.39 &33.17 &32.48 &\textcolor{secondgreen}{37.87}  \\
        
        \toprule
        \toprule

        \multicolumn{11}{c}{\textbf{(B) Prior Knowledge-based Question-Answering}} \\
        \midrule
        Task & Dataset &Random &V-Chat &V-ChatGPT &Otter &PandaGPT &Valley &mPLUG &V-LLaMA &C-UniVi \\
        \midrule
        \multirow{3}{*}{Prior Knowledge} &TV-QA* &20.00 &26.15 &\textcolor{blue}{\textbf{28.76}} &27.65 &\textcolor{secondgreen}{27.85} &23.70  &23.95 &24.75 &23.05\\
                                &MV-QA* &26.15 &34.11 &\textcolor{secondgreen}{36.52}  &\textcolor{blue}{\textbf{37.06}} &\textcolor{blue}{\textbf{37.06}} &32.59 &30.17 &32.41 &33.57 \\
                                &NBA-QA* &27.26 &28.57 &22.45 &\textcolor{blue}{\textbf{34.26}} &31.05 &\textcolor{secondgreen}{31.34} &25.07 &26.24 &25.66 \\
        \midrule
        \multicolumn{2}{c|}{Average Score} &24.47 &29.61 &29.24 &\textcolor{blue}{\textbf{32.99}} &\textcolor{secondgreen}{31.99} &29.21 &26.40 &27.80 &27.43 \\
                                
        \toprule
        \toprule

        \multicolumn{11}{c}{\textbf{(C) Comprehension and Decision-Making}} \\
        \midrule
        Task & Dataset &Random &V-Chat &V-ChatGPT &Otter &PandaGPT &Valley &mPLUG &V-LLaMA &C-UniVi \\
        \midrule
        \multirow{2}{*}{Driving} &License Exam*$^{\dagger}$ &36.81 &38.89 &\textcolor{secondgreen}{41.67} &\textcolor{blue}{\textbf{52.78}} &\textcolor{secondgreen}{41.67} &\textcolor{secondgreen}{41.67} &33.34 &30.56 &38.89 \\
                                &Decision-Making* &44.21 &55.38 &\textcolor{blue}{\textbf{58.21}} &48.72 &56.03 &\textcolor{secondgreen}{56.54} &51.03 &49.10 &53.08 \\ 
        \midrule
        3D Scene &SQA3D* &25.00 &31.42 &\textcolor{blue}{\textbf{37.20}} &29.65 &30.76 &\textcolor{secondgreen}{33.30} &32.00 &31.15 &29.07\\
        \midrule
        \multicolumn{2}{c|}{Average Score} &35.04 &42.50 &\textcolor{blue}{\textbf{46.50}} &41.90 &43.05 &\textcolor{secondgreen}{44.27} &39.88 &38.21 &40.64  \\

        \toprule
        \toprule

        \multicolumn{11}{c}{\textbf{(D) Final Result}} \\
        \midrule
         \multicolumn{2}{c|}{}  &Random &V-Chat &V-ChatGPT &Otter &PandaGPT &Valley &mPLUG &V-LLaMA &C-UniVi \\
        \midrule        
        \multicolumn{2}{c|}{Average Score} &28.45 &35.41 &\textcolor{blue}{\textbf{38.52}} &37.47 &\textcolor{secondgreen}{37.52} &33.96 &33.15 &32.83 &35.31  \\
        
        \bottomrule[1pt]        
        \end{tabular}
        }}
\end{table*}

\subsection{Comprehension and Decision-making}
\label{sec:comprehension}
Humans possess the innate ability to comprehend complex scenarios and make informed decisions and judgments. As shown in Fig.~\ref{fig:visual_ability} (C), to assess a similar capability in Video-LLMs, we propose evaluations in the realms of 3D scene understanding and autonomous-driving related tasks.

\vspace{0.1cm}
\noindent\textbf{3D Scene Comprehension.} Indoor scene comprehension and navigation hold significant practical implications. The complexity arises from the necessity for extensive knowledge-intensive reasoning to understand different situations (scenes and locations). The SQA3D dataset~\cite{ma2022sqa3d} is introduced to evaluate the 3D scene comprehension of Video-LLMs within the video modality. The models are tasked with understanding their environment and engaging in perception, reasoning, and action to accomplish the task.

\vspace{0.1cm}
\noindent\textbf{Driver’s License Examination.} Video-based questions in driver's license examinations assess the ability of candidates to interpret simple animations depicting motor vehicle and driver status, requiring judgments of potential anomalies. In this task, we challenge Video-LLMs to comprehend scenarios and answer exam questions.

\vspace{0.1cm}
\noindent\textbf{Driving Decision-Making.} Making decisions for real-world driving scenarios is a more intricate task that demands a higher level of scene understanding and decision-making ability. For this task, we compile a diverse collection of YouTube driving videos depicting complex traffic situations and accidents. We conduct manual annotations for scene analysis and accident causes. Our expectation is that the model can effectively comprehend the origins of these complex traffic situations or accidents and make correct decisions to prevent their occurrence.

\subsection{Automatic Evaluation Toolkit}
\label{sec:toolkit}
LLMs are known for generating long-form text responses, often without adhering to a fixed format, making it challenging to quantify the correctness of their answers. To address this, we propose an automatic evaluation toolkit to systematically assess the performance of Video-LLMs. 
Our toolkit provides three metrics to map the output of Video-LLMs to pre-defined answer choices and subsequently calculating the final scores. The first one is Probability~\cite{hendrycks2020measuring}, a logits-based metric to acquire the probability of the next token following the prompt and treat the highest probability option as the prediction:
\begin{equation}
    \text { Choice }=\arg \max _{i \in\left\{A, B, C, D,  \ldots\right\}} P\left(\text{Token}_i\mid\text{Prompt}\right).
\end{equation}
The other two metrics are sentence-based, leveraging the natural language understanding capabilities of LLMs to obtain options. T5-based~\cite{raffel2020exploring} one calculates the textual similarities of generated sequences and options. GPT-3.5-based~\cite{ouyang2022training} transforms the sequences to a fixed format with prompt \textit{`Please output your responses in the form of a dictionary {"maximum probability":"xxx"}, where xxx is A or B or C or ...'}.
All the above metrics can be implemented automatically with our toolkit, and users can analysis  the ability of video-LLMs to comprehend video content and provide accurate responses to questions faithfully.

\begin{figure}[t]
    \centering
    \includegraphics[width=1\columnwidth]{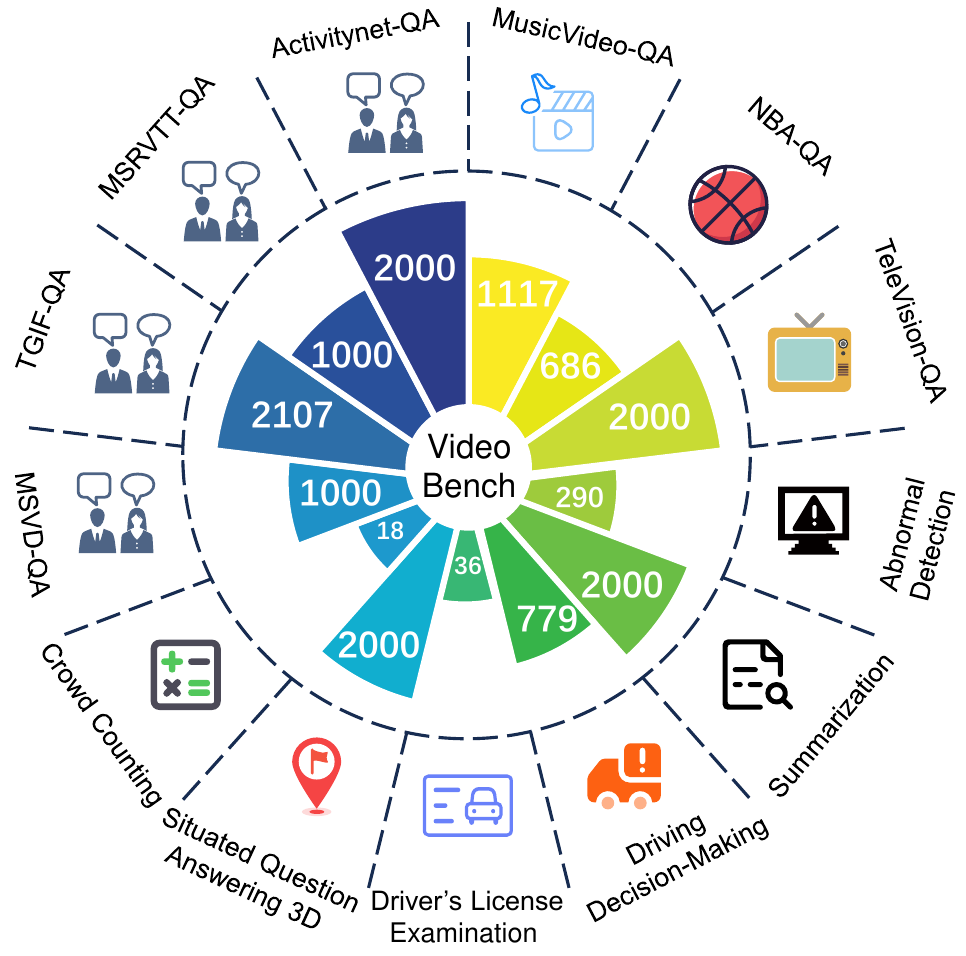}
    \caption{The detailed amount of QA pairs of different tasks.}
    \label{fig:pie}
\end{figure}

\section{Experiment and Result}

\noindent\textbf{Implementation details.}
\label{sec:imp_detail}
The detailed statistics of \textit{Video-Bench} are listed in Fig.~\ref{fig:pie}. To mitigate the impact of randomness, we multiply an additional weight of 0.5 for tasks with a smaller quantity of questions during the computation of the final average score.
To ensure a fair comparison, we utilize the 7B LLM backend versions for all tested Video-LLMs during the inference process, thereby mitigating language ability discrepancy stemming from different model sizes. The GPT-based metric are employed in the reported results by default, and the API version is set to \textit{gpt-3.5-turbo-0613} in the automatic evaluation toolkit.

\vspace{0.1cm}
\noindent\textbf{Results on Video-exclusive Understanding.}
\label{sec:result_exclusive}
To evaluate the video-exclusive understanding ability, we validate Video-LLMs on the traditional basic QA tasks, summarization, abnormal detection and crowd counting tasks, as reported in Table.~\ref{table:main_result} (A). We have three observations. (\textit{i}) Most Video-LLMs perform well on the four traiditional QA datasets due to the simplicity of their questions, especially the Video-ChatGPT~\cite{maaz2023video} and Otter~\cite{li2023otter} with massive video instruction data, and the PandaGPT~\cite{su2023pandagpt} with a well-pretrained video encoder from ImageBind~\cite{girdhar2023imagebind}, which suggests extending the video data scale could be effective. (\textit{ii}) Existing Video-LLMs are not temporal-sensitive. They cannot effectively summarize the order of each operation in YouCook2, and cannot respond effectively on the timestamp-related problems in UCF-Crime. 
(\textit{iii}) These methods almost fail in the crowd counting task. These failure may come from the weak ability of precise locating and the temporal association.

\begin{figure*}[ht]
  \centering

  \begin{minipage}{.25\textwidth}
    \centering
    \includegraphics[width=\linewidth]{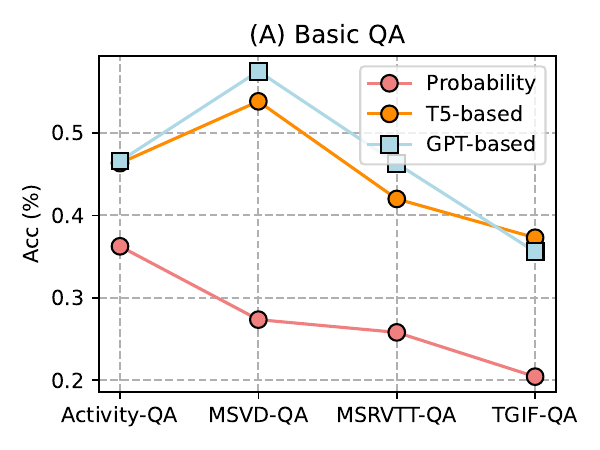}
  \end{minipage}%
  \begin{minipage}{.25\textwidth}
    \centering
    \includegraphics[width=\linewidth]{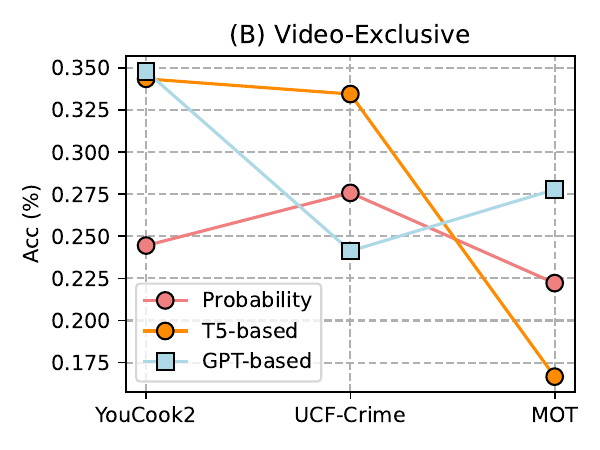}
  \end{minipage}%
  \begin{minipage}{.25\textwidth}
    \centering
    \includegraphics[width=\linewidth]{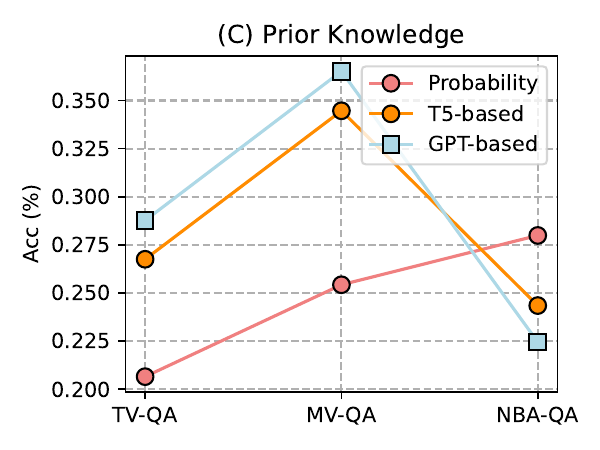}
  \end{minipage}%
  \begin{minipage}{.25\textwidth}
    \centering
    \includegraphics[width=\linewidth]{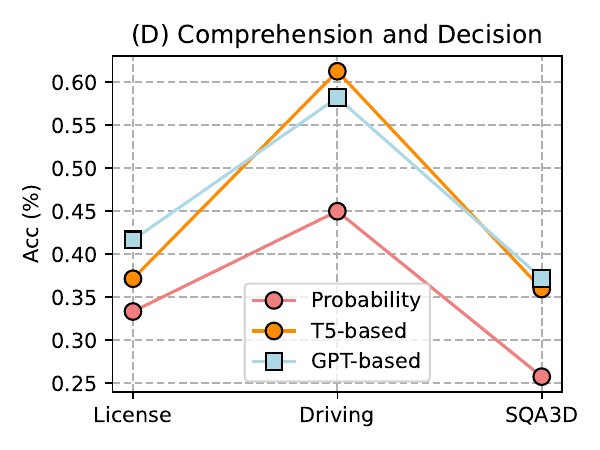}
  \end{minipage}

  \hspace{0.1\textwidth} 
  \begin{minipage}[t]{1\textwidth}
    \centering

    \setlength\tabcolsep{1.15mm}
    \renewcommand\arraystretch{1.0}
    {
    \small 
    \scalebox{0.95}{
    \begin{tabular}{l|cccc|ccc|ccc|ccc}
        \toprule[1pt]
        \multirow{2}{*}{\textbf{Metric}} &\multicolumn{4}{c|}{\textbf{Basic QA}}  &\multicolumn{3}{c|}{\textbf{Video-Exclusive}}  &\multicolumn{3}{c|}{\textbf{Prior Knowledge}}  &\multicolumn{3}{c}{\textbf{Comprehension and Decision}}  \\
         &Activity &MSVD &MSRVTT &TGIF &YouCook2  &Crime &MOT &TVQA  &MVQA &NBAQA &License  &Driving &SQA3D\\
        \midrule [1pt]
        Probability~\cite{hendrycks2020measuring} &36.25 &27.35 &25.80 &20.43 &24.45 &27.58 &\underline{22.22} &20.65 &25.43 &\textbf{27.99} &33.33 &45.00 &25.75\\ 
        T5-based~\cite{raffel2020exploring} &\underline{46.35} &\underline{53.85} &\underline{42.00} &\textbf{37.29} &\underline{34.35} &\textbf{33.45} &16.67 &\underline{26.75} &\underline{34.47} &\underline{24.34} &\underline{37.14} &\textbf{61.28} &\underline{35.95} \\
        GPT-based~\cite{ouyang2022training}  & \textbf{46.60} &\textbf{57.50} &\textbf{46.30} &\underline{35.59} &\textbf{34.80} &\underline{24.13} &\textbf{27.77} &\textbf{28.76} &\textbf{36.52} &22.45 &\textbf{41.67} &\underline{58.21} &\textbf{37.20}\\

        \bottomrule[1pt]
        \end{tabular}}
        }
  \end{minipage}
  
  \caption{Comparison results of different metrics of Video-ChatGPT~\cite{maaz2023video} on all datasets.}
  \label{fig:metrics}
\end{figure*}

\noindent\textbf{Results on Prior Knowledge-based QA.}
\label{sec:result_prior}
Compared to enormous training data of LLMs, existing Video-LLMs are trained with limited instruction tuning data as Table.~\ref{table:vllm}, resulting in the poor ability to recognize objects and information in specific domains. As shown in Table.~\ref{table:main_result} (B), we can have two observations. (\textit{i}) Existing methods lack visual prior knowledge, which means they struggle to establish effective connection between the video and knowledge. For example, in NBA-QA task, even the players and technical actions are stored in the LLM backend, they cannot answer the questions when watching videos. Otter~\cite{li2023otter}, which has the most instruction tuning data, achieves the best performance in this project, indicating that some prior knowledge is indeed contained in MIMIC-IT. (\textit{ii}) Their poor performance on MV-QA indicates that they have limited audio understanding ability, since only some of the Video-LLMs possess audio modules. PandaGPT~\cite{su2023pandagpt} with the audio module of ImageBind shows the consistent results with the champion Otter~\cite{li2023otter} in MV-QA, proving that adding an audio encoder might improve this problem. In conclusion, existing Video-LLMs are requiring abundant prior knowledge pre-training for general domains on different modalities.

\vspace{0.1cm}
\noindent\textbf{Results on Comprehension and Decision-making.}
\label{sec:result_decision}
The performance of existing Video-LLMs on 3D scene understanding and driving decision-making tasks is shown in Table.~\ref{table:main_result} (C). In these tasks, Video-ChatGPT~\cite{maaz2023video} continues to perform the best, thanks to its robust video instruction tuning. The followings are the Valley~\cite{luo2023valley}, which also possess powerful multi-modal understanding ability from vast instruct-tuning videos. To enhance the comprehensive and decision-making abilities, we suggest that future Video-LLMs must be trained with more prior knowledge and larger-scale data to cover more diverse domains. Besides, adopting Reinforcement Learning from Human Feedback (RLHF) and larger model capability is also important for generalization and specific applications.

\vspace{0.1cm}
\noindent\textbf{Results on Different Metrics.}
Our \textit{Video-Bench} consists of a series of multiple-choice questions. Compared to open-ended questions, this test is relatively straightforward. However, due to the uncertainty and free form of LLM outputs, there is still room for designing more robust metrics. We evaluate the results of the best tested model, comparing the results with Probability~\cite{hendrycks2020measuring}, T5-based and the GPT-based metrics. as shown in the Fig.~\ref{fig:metrics}. It can be seen that the result of Probability is overall low, because the output of Video-LLMs cannot effectively give a clear choice answer and the probability-based mapping may not faithfully reflect the correctness. Therefore, we recommend adopting GPT as the metric, especially considering the Video-LLMs with fewer LLM parameters and unstable outputs.

\begin{figure*}
    \centering
    \includegraphics[width=2\columnwidth]{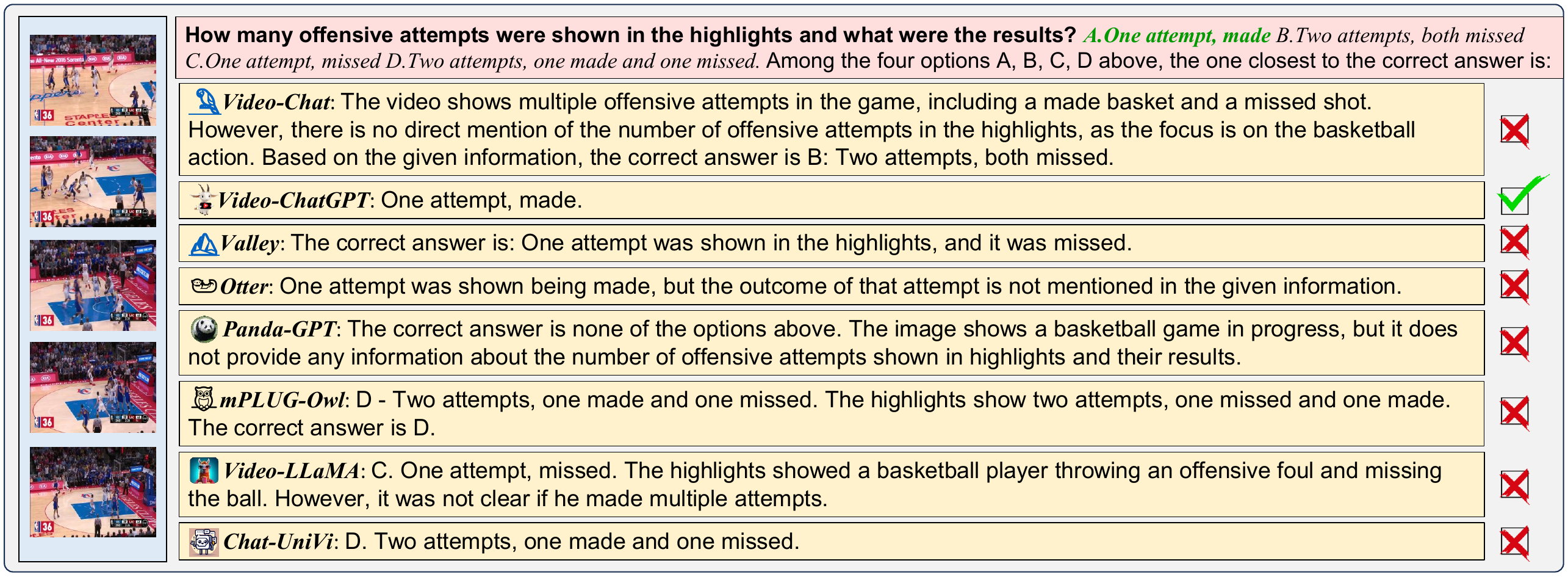}
    \caption{The illustrative sample of the generated responses from different Video-LLMs.}
    \label{fig:visual_llm}
    \vspace{-0.1cm}
\end{figure*}

\section{Visualization and Multi-Dimension Analysis}

\noindent\textbf{Visualization.}
\label{sec:visualization}
Fig.~\ref{fig:visual_llm} illustrates a set of typical responses from tested Video-LLMs. It can be observed that only Video-ChatGPT~\cite{maaz2023video} provides the correct response, while other models engage in discussions related to the video but fail to make the correct judgment after a lengthy discourse. This highlights the issue that the models struggle with questions
with even the most fundamental prior knowledge. This situation reflects the current state of Video-LLMs, which can generate responses related to videos while lacking trustful reference value. 
Therefore, we can conclude that current Video-LLMs are limited to generating human-like text while lacking the desired intelligence.

\begin{figure}[ht]
    \vspace{-0.2cm}
    \centering
    \includegraphics[width=1\columnwidth]{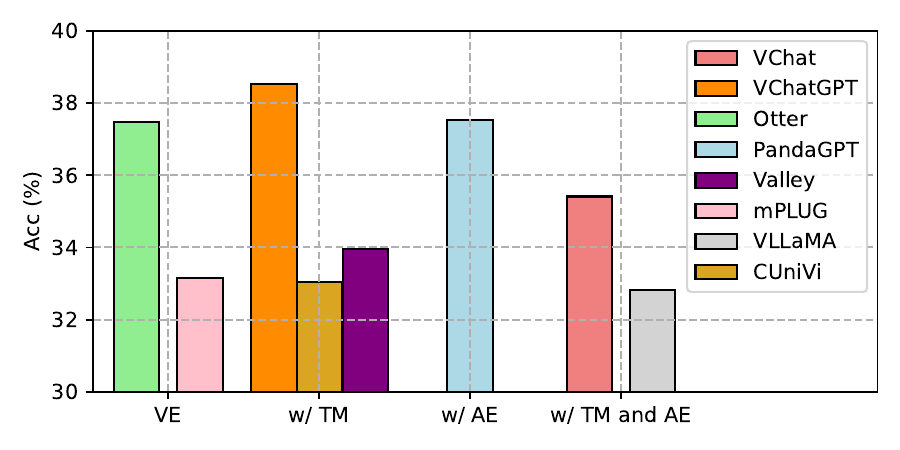}
    \caption{The impact of different module configuration. The w/ TM, w/ AE, and w/ TM and AE means the model contains Temporal Module (TM) or Audio Encoder (AE) or both of them.}    
    \label{fig:impact_module}
    \vspace{-0.1cm}
\end{figure}

\vspace{0.1cm}
\noindent\textbf{Multi-dimension Analysis.}
\label{sec:multi_dim_analysis}
In Fig.~\ref{fig:impact_module}, a comparative analysis of Video-LLMs with different modules is presented. We can conclude that with the current data and training setting, Video-LLMs lack tailored focus on  the three-level ability of video comprehension. And the empirically proposed modules  have not yielded significant improvements. 

\begin{figure}
  \centering

  \begin{minipage}{.25\textwidth}
    \centering
    \includegraphics[width=\linewidth]{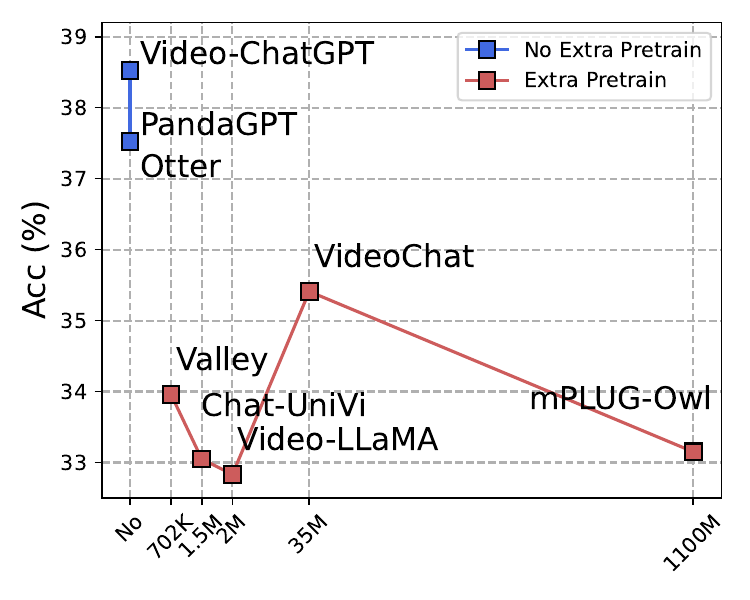}
  \end{minipage}%
  \begin{minipage}{.25\textwidth}
    \centering
    \includegraphics[width=\linewidth]{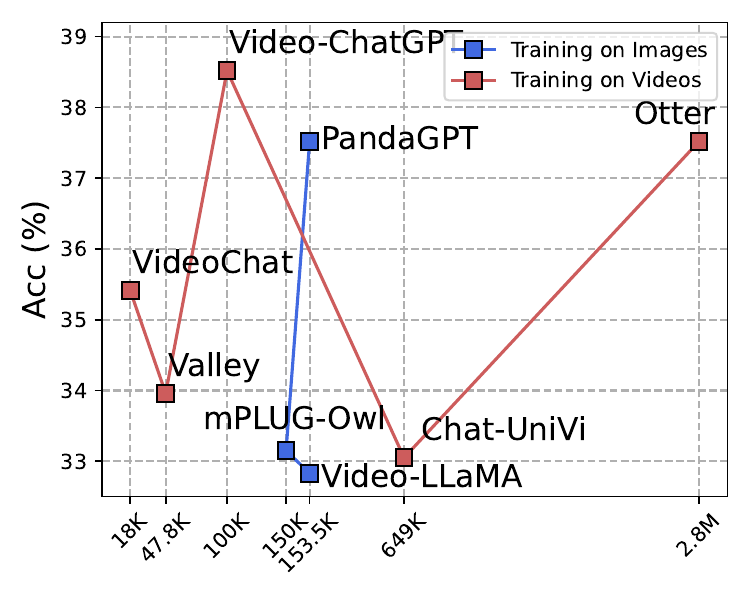}
  \end{minipage}%
  \caption{Impact of different datasize of pretrain data (left) or instruction tuning data (right).}
  \label{fig:impact_datasize}
  \vspace{-0.2cm}
\end{figure}

We also analysis the impact of different data sizes in pre-training or instruction tuning process, as shown in Fig.~\ref{fig:impact_datasize}. It can be observed that pre-training datasize may not necessarily play a decisive role, as the top-3 models, Video-ChatGPT~\cite{maaz2023video}, PandaGPT~\cite{su2023pandagpt} and Otter~\cite{li2023otter}, have no extra pretraining process. We suppose that the video encoders have received adequate training in multimodal pre-training.
In contrary, the influence of the instruction tuning datasize is notably evident, showing two trends: (\textit{i}) The models trained on videos demonstrate overall better performance compared to those trained on images. This substantiates that native video data facilitates enhanced comprehension of video information by Video-LLMs.
(\textit{ii}) Model performance is positively correlated with the amount of video instruction tuning data. Video-ChatGPT~\cite{maaz2023video} and Otter~\cite{li2023otter} trained on large-scale video instruction tuning datasets are significantly better than other models.

\section{Discussion and Conclusion}
According to the above experimental results, we can conclude that the existing models are far from the truly intelligent Video-LLM that can fully understand the visual and audio content in videos, and help people precisely summarize videos, explain details with priority knowledge or help providing global perception and making decisions. We believe there are primarily three improvement directions.

\vspace{0.1cm}
\noindent\textbf{Vision Encoder with Temporal Awareness.} Existing methods process videos as frame clips, potentially missing crucial temporal information. Ideal Video-LLMs should understand the temporal sequence, possibly by selectively choosing keyframes or sampling frames to traverse the content efficiently.

\vspace{0.1cm}
\noindent\textbf{Domain-Specific Prior Knowledge Pre-training.} Lack of visual prior knowledge hinders accurate video comprehension. Incorporating domain-specific prior knowledge through pre-training can enhance domain expertise.

\vspace{0.1cm}
\noindent\textbf{Long Video Understanding.} One key differentiation point of Video-LLMs when compared to Image-LLMs should be the capability of processing long videos, which is highly neglected by existing research. Due to the memory and computation constraint, how to efficiently compress the past frames and design an effective memory mechanism is super crucial. 

Simultaneously, we also require more robust and effective evaluation metrics that can measure the long-text response of Video-LLMs. The utilization of GPT or similar LLMs in this process could be highly beneficial.

\clearpage
{
    \small
    \bibliographystyle{ieeenat_fullname}
    \bibliography{main}
}

\clearpage
\setcounter{page}{1}
\maketitlesupplementary

\begin{table*}[!htpb]
    \vspace{-3cm}
    \caption{\textbf{T5-large experiment results of tested video-LLMs on various tasks.} `*' denotes the QA-pairs are re-constructed or annotated by \textit{Video-Bench}. `$^{\dagger}$' denotes the tasks with fewer videos and multiplying the weight by 0.5 when calculating the final result. For each task, \textcolor{blue}{\textbf{blue}} and \textcolor{secondgreen}{green} mark the \textcolor{blue}{\textbf{first}} and \textcolor{secondgreen}{second} place respectively. All the reported results are accuracy (\%) calculated with GPT-3.5-based~\cite{ouyang2022training} metric. The ``Video-'' and ``Chat-'' are abbreviated to ``V-'' and ``C-''.}
    
    \label{table:t5}
    \centering
    \setlength\tabcolsep{1.3mm}
    \renewcommand\arraystretch{1}
    {
    \small 
    \scalebox{1.0}{
    \begin{tabular}{l|l|ccccccccc}
        \toprule[1pt]
        \multicolumn{11}{c}{\textbf{(A) Video-Exclusive Understanding}} \\
        \midrule
        Task & Dataset &Random &V-Chat &V-ChatGPT &Otter &PandaGPT &Valley &mPLUG &V-LLaMA &C-UniVi \\
        \midrule
        \multirow{4}{*}{Basic QA} &Activitynet-QA &34.58 &\textcolor{secondgreen}{46.7} &46.35 &44.15 &46.1 &37.65 &42.3 &44.05 &\textcolor{blue}{\textbf{47.95}} \\
                                  &MSVD-QA* &26.22 &43.25 &\textcolor{blue}{\textbf{53.85}} &\textcolor{secondgreen}{53.35} &43.65 &30.93 &42.6 &42.7 &44.65  \\
                                  &MSRVTT-QA*  &26.50  &37.0 &\textcolor{secondgreen}{42.0} &\textcolor{blue}{\textbf{42.9}} &39.4 &28.48 &34.65 &34.75 &41.25 \\
                                  &TGIF-QA &22.37 &{36.02} &\textcolor{secondgreen}{37.29} &35.41 &34.17 &33.84 &32.75 &34.27 &\textcolor{blue}{\textbf{42.30}} \\
        \midrule
        Summarization & YouCook2*  &25.00 &30.0 &\textcolor{blue}{\textbf{34.35}} &30.4 &\textcolor{secondgreen}{31.3} &26.95 &27.25 &28.4 &30.65 \\
        \midrule
        Abnormal Detection & UCF-Cirme* &16.67 &18.62 &\textcolor{blue}{\textbf{33.45}} &{26.21} &24.83 &13.45 &18.62 &22.07 &\textcolor{secondgreen}{30.86} \\
        \midrule
        Crowd Counting & MOT*$^{\dagger}$  &16.67 &\textcolor{secondgreen}{22.22}   &16.67  &\textcolor{blue}{\textbf{27.78}} &5.56 &11.11  &11.10  &16.67  &11.11 \\
        \midrule
        \multicolumn{2}{c}{Average Score}  &25.85   &34.26  &\textcolor{blue}{\textbf{39.33}}   &\textcolor{secondgreen}{37.89} &34.19  &27.21  &31.34  &33.01  & 37.42        \\
        
        \toprule
        \toprule

        \multicolumn{11}{c}{\textbf{(B) Prior Knowledge-based Question-Answering}} \\
        \midrule
        Task & Dataset &Random &V-Chat &V-ChatGPT &Otter &PandaGPT &Valley &mPLUG &V-LLaMA &C-UniVi  \\
        \midrule
        \multirow{3}{*}{Prior Knowledge} &TV-QA* &20.00 &\textcolor{blue}{\textbf{28.3}} &26.75 &25.0 &\textcolor{secondgreen}{27.55} &22.15 &24.25 &25.45 &23.05 \\
                                &MV-QA*  &26.15 &30.26 &34.47 &32.41 &\textcolor{blue}{\textbf{34.91}} &27.13 &29.01 &27.84 &\textcolor{secondgreen}{33.48} \\
                                &NBA-QA* &27.26 &25.36 &24.34 &\textcolor{blue}{\textbf{32.51}} &26.53 &25.36 &26.82 &\textcolor{secondgreen}{28.13} &24.49  \\
        \midrule
        \multicolumn{2}{c}{Average Score}  &24.47   &27.97  &28.52  &\textcolor{blue}{\textbf{29.97}}   &\textcolor{secondgreen}{29.66}     &24.88  &26.69  &27.14  &27.01 \\
                                
        \toprule
        \toprule

        \multicolumn{11}{c}{\textbf{(C) Comprehension and Decision-Making}} \\
        \midrule
        Task & Dataset &Random &V-Chat &V-ChatGPT &Otter &PandaGPT &Valley &mPLUG &V-LLaMA &C-UniVi  \\
        \midrule
        \multirow{2}{*}{Driving} &License Exam*$^{\dagger}$ &36.81 &25.0 &{37.14} &\textcolor{blue}{\textbf{55.56}} &36.11 &30.56 &36.11 &25.0 &\textcolor{secondgreen}{50.0} \\
                                &Decision-Making* &44.21 &60.77 &\textcolor{secondgreen}{61.28} &47.44 &\textcolor{blue}{\textbf{62.18}} &56.28 &53.21 &49.49 &49.74 \\ 
        \midrule
        3D Scene &SQA3D*  &25.00 &30.08 & \textcolor{blue}{\textbf{35.95}} & 27.45& 30.25&\textcolor{secondgreen}{35.65} & 32.35& 30.5& 27.4  \\
        \midrule
        \multicolumn{2}{c}{Average Score}  &35.04   &41.34  &\textcolor{blue}{\textbf{46.32}}   &41.07  & \textcolor{secondgreen}{44.19}    &42.88  &41.45  &37.00  &40.86   \\

        \toprule
        \toprule

        \multicolumn{11}{c}{\textbf{(D) Final Result}} \\
        \midrule
        \multicolumn{2}{c|}{Task} &Random &V-Chat &V-ChatGPT &Otter &PandaGPT &Valley &mPLUG &V-LLaMA &C-UniVi \\
        \midrule        
        \multicolumn{2}{c}{Average Score} 
        &28.45  &34.53  &\textcolor{blue}{\textbf{38.06}}   &\textcolor{secondgreen}{36.31}     &36.02  &31.66  &33.16  &32.38  &35.10 \\

        \toprule
        \toprule

        \multicolumn{11}{c}{\textbf{(E) Comparison Result of GPT-based}} \\
        \midrule
         \multicolumn{2}{c|}{Task}  &Random &V-Chat &V-ChatGPT &Otter &PandaGPT &Valley &mPLUG &V-LLaMA &C-UniVi \\
        \midrule        
        \multicolumn{2}{c|}{Video-Exclusive Understanding} &25.85 &34.12 &\textcolor{blue}{\textbf{39.82}} &37.52 &\textcolor{secondgreen}{37.54} &28.39 &33.17 &32.48 &37.87  \\
        \multicolumn{2}{c|}{Prior Knowledge-based QA} &24.47 &29.61 &29.24 &\textcolor{blue}{\textbf{32.99}} &\textcolor{secondgreen}{31.99} &29.21 &26.40 &27.80 &27.43  \\
        \multicolumn{2}{c|}{Comprehension and Decision} &35.04 &42.50 &\textcolor{blue}{\textbf{46.50}} &41.90 &43.05 &\textcolor{secondgreen}{44.27} &39.88 &38.21 &40.64  \\
        \midrule
        \multicolumn{2}{c|}{Average Score} &28.45 &35.41 &\textcolor{blue}{\textbf{38.52}} &37.47 &\textcolor{secondgreen}{37.52} &33.96 &33.15 &32.83 &35.31  \\
        
        \bottomrule[1pt]        
         \end{tabular}
        }}

\end{table*}

\section{T5 evaluation}
In our answer evaluation benchmark project, we explore two approaches: GPT-based metric and T5-based metric. T5-based metric serves as an auxiliary tool in the evaluation process, offering advantages in terms of cost, deployment, and performance. It provides a cost-effective solution by eliminating the need for ChatGPT API usage and allows for offline deployment on personal servers. As shown in Table~\ref{table:t5}, T5-based results demonstrate comparable performance to GPT-based in answer evaluation tasks, making it a valuable addition to our benchmark project for reliable and efficient assessment.

\section{Visualization Samples}
In this part, we provide more samples of on all datasets concluded in \textit{Video-Bench}, to illustrate the performance and behaviour of the tested Video-LLMs.

\subsection{Video-exclusive Understanding}

\begin{figure*}
    \centering
    \includegraphics[width=2\columnwidth]{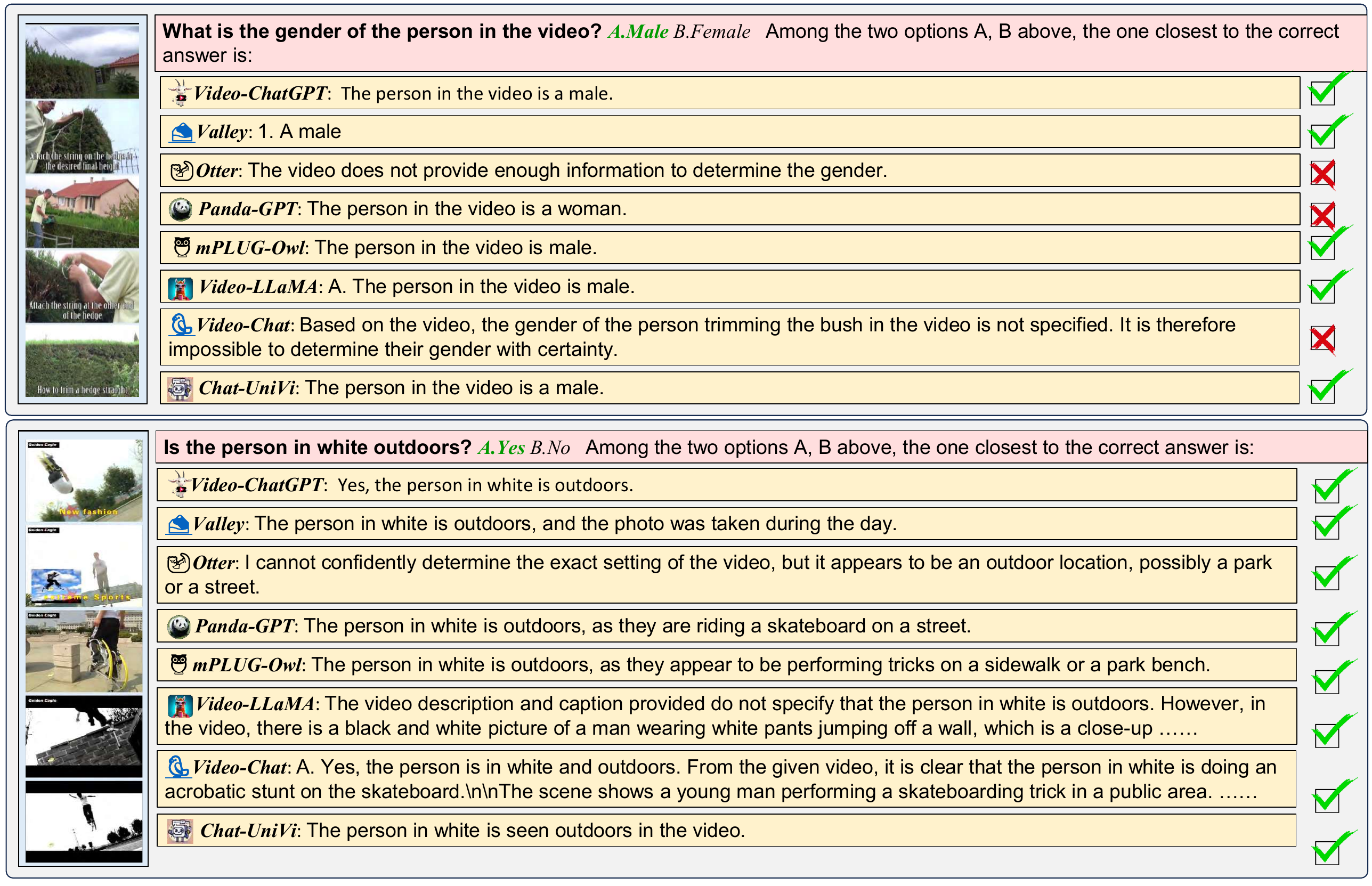}
    \caption{The illustrative sample of the generated responses from different Video-LLMs for Activitynet-QA.}
    \label{fig:sup_activity}
\end{figure*}

\vspace{0.1cm}
\noindent\textbf{Activitynet-QA.} The results of the Activitynet-QA is shown in  Fig.~\ref{fig:sup_activity}. As mentioned in Sec~\ref{sec:result_exclusive}, Video-LLMs perform well on these simple questions. The similar results are shown on the remaining three datasets of \textit{Basic QA}.

\begin{figure*}
    \centering
    \includegraphics[width=2\columnwidth]{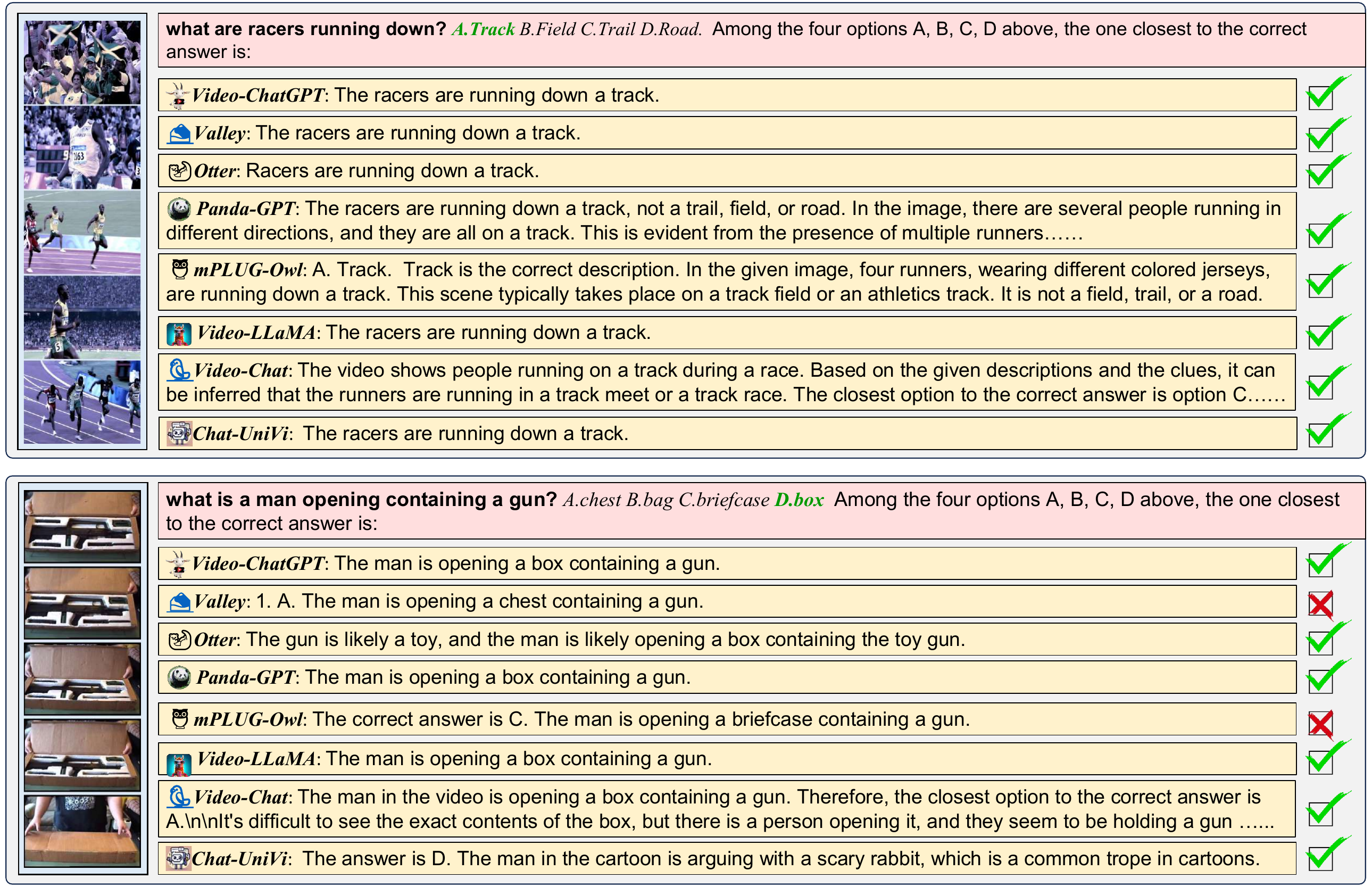}
    \caption{The illustrative sample of the generated responses from different Video-LLMs for MSVD-QA.}
    \label{fig:sup_msvd}
\end{figure*}

\vspace{0.1cm}
\noindent\textbf{MSVD-QA.} The results of the MSVD-QA is shown in  Fig.~\ref{fig:sup_msvd}. As part of the \textit{Basic QA}, the performance of Video-LLMs here are overall good.

\begin{figure*}
    \centering
    \includegraphics[width=2\columnwidth]{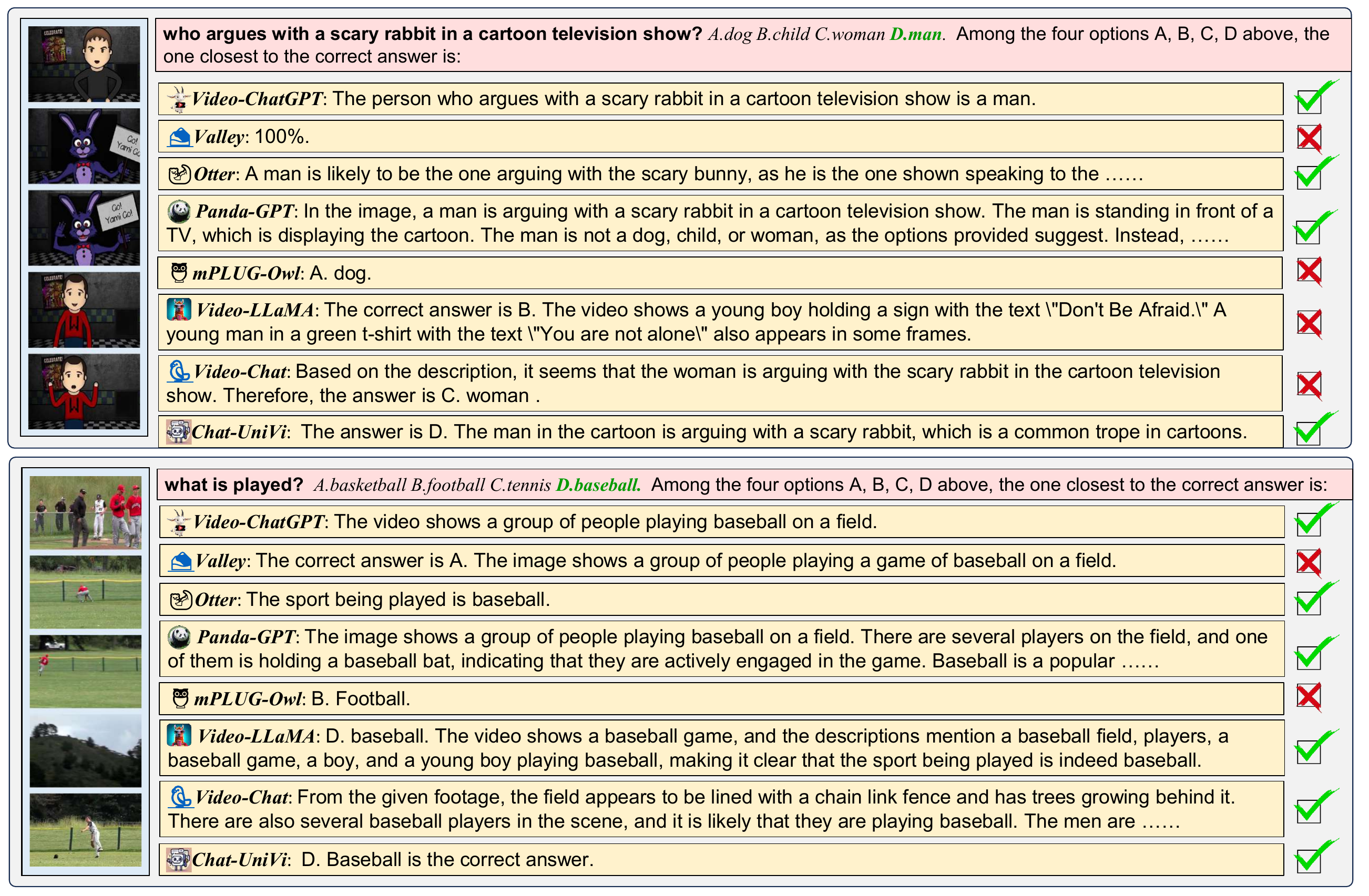}
    \caption{The illustrative sample of the generated responses from different Video-LLMs for MSRVTT-QA.}
    \label{fig:sup_msrvtt}
\end{figure*}

\vspace{0.1cm}
\noindent\textbf{MSRVTT-QA.} The results of the MSRVTT-QA is shown in  Fig.~\ref{fig:sup_msrvtt}. The results shows a similar trend of the above.

\begin{figure*}
    \centering
    \includegraphics[width=2\columnwidth]{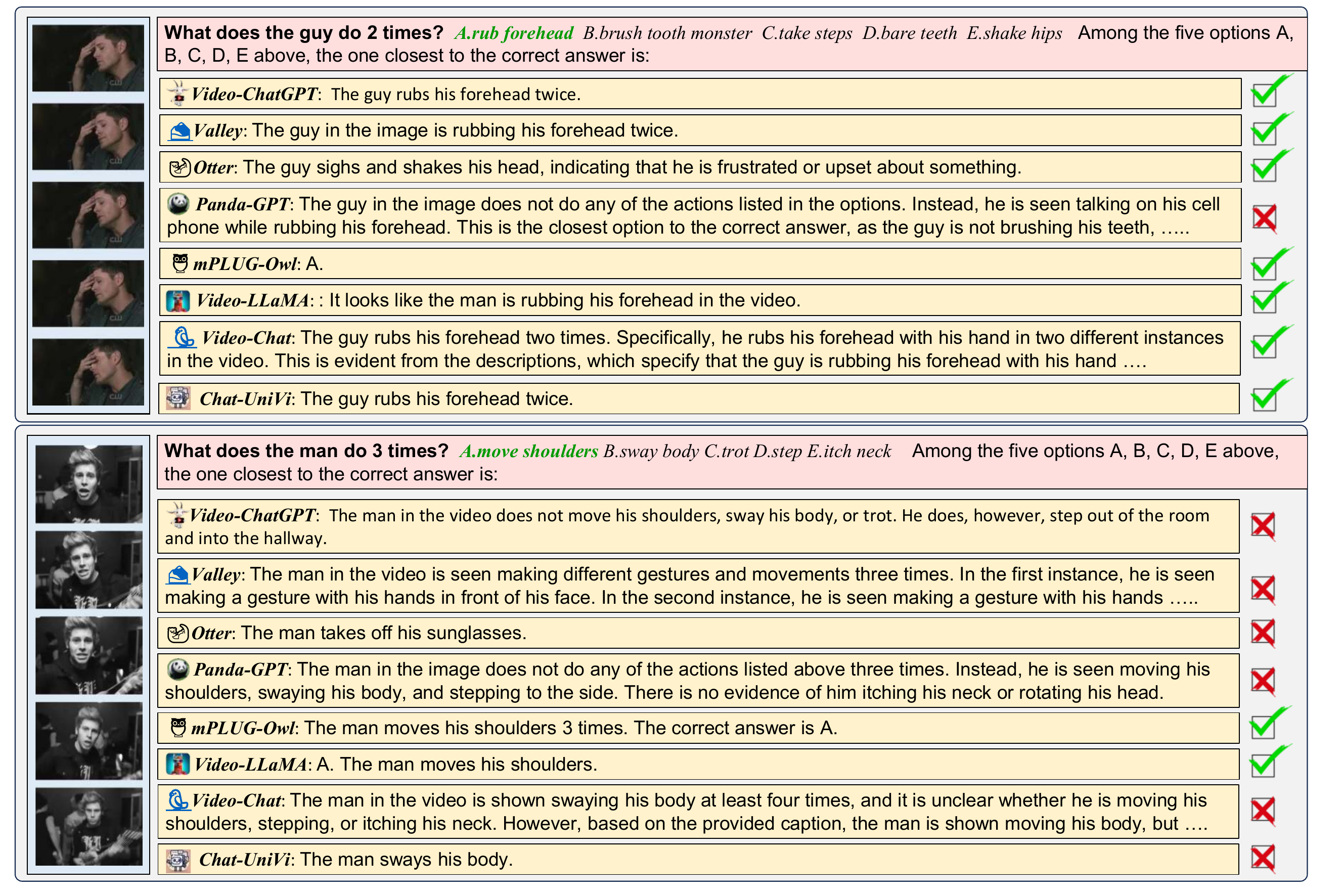}
    \caption{The illustrative sample of the generated responses from different Video-LLMs for TGIF-QA.}
    \label{fig:sup_tgif}
\end{figure*}

\vspace{0.1cm}
\noindent\textbf{TGIF-QA.} The results of the TGIF-QA is shown in  Fig.~\ref{fig:sup_tgif}. Results prove that Video-LLMS can also understand simple GIFs.

\begin{figure*}
    \centering
    \includegraphics[width=2\columnwidth]{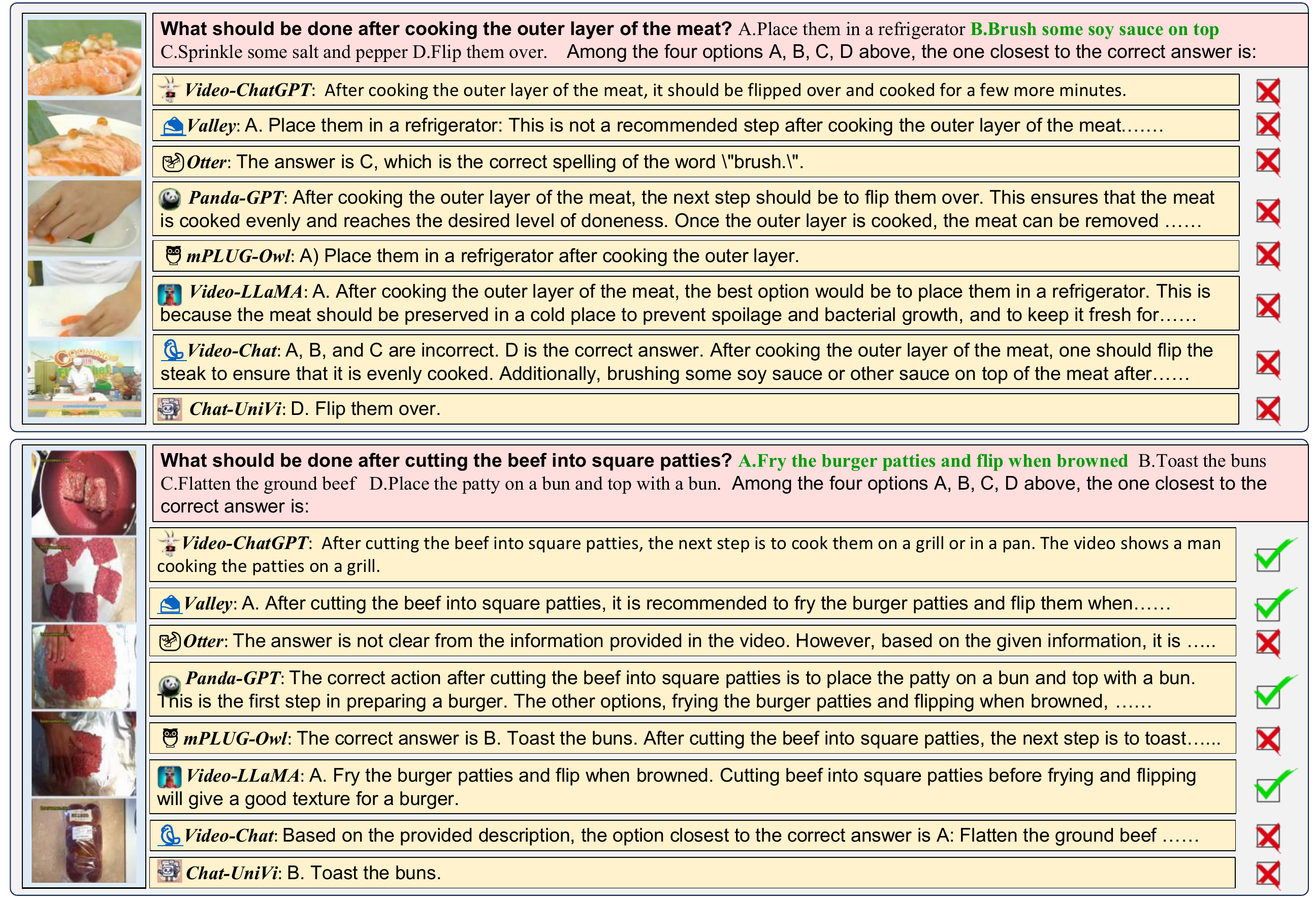}
    \caption{The illustrative sample of the generated responses from different Video-LLMs for YouCook2.}
    \label{fig:sup_youcook}
\end{figure*}

\vspace{0.1cm}
\noindent\textbf{YouCook2.} The results of the YouCook2 is shown in  Fig.~\ref{fig:sup_youcook}. The poor results show that existing Video-LLMs possess limited temporal awareness, and they are difficult to summarize the sequence of action steps.

\begin{figure*}
  \centering
  \includegraphics[width=2\columnwidth]{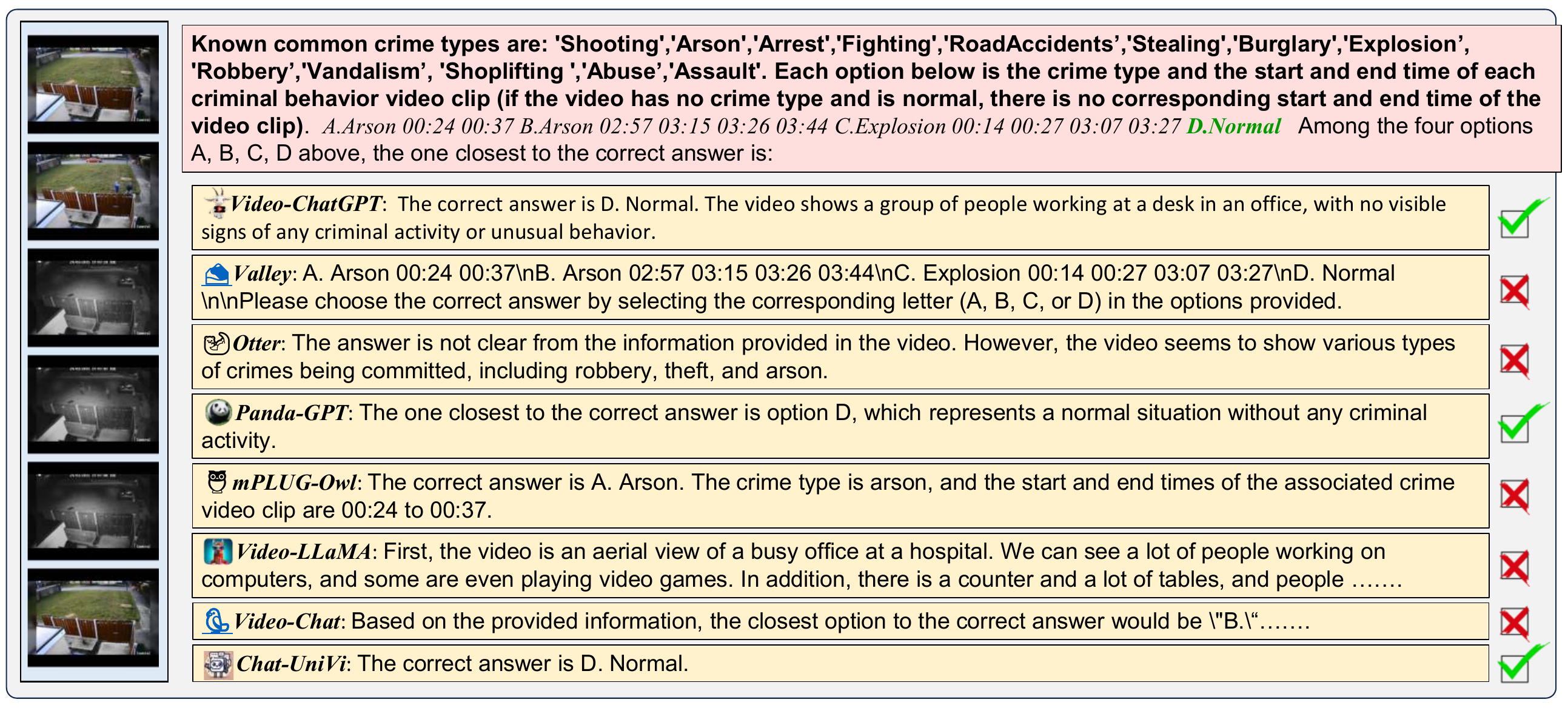}
  
  \includegraphics[width=2\columnwidth]{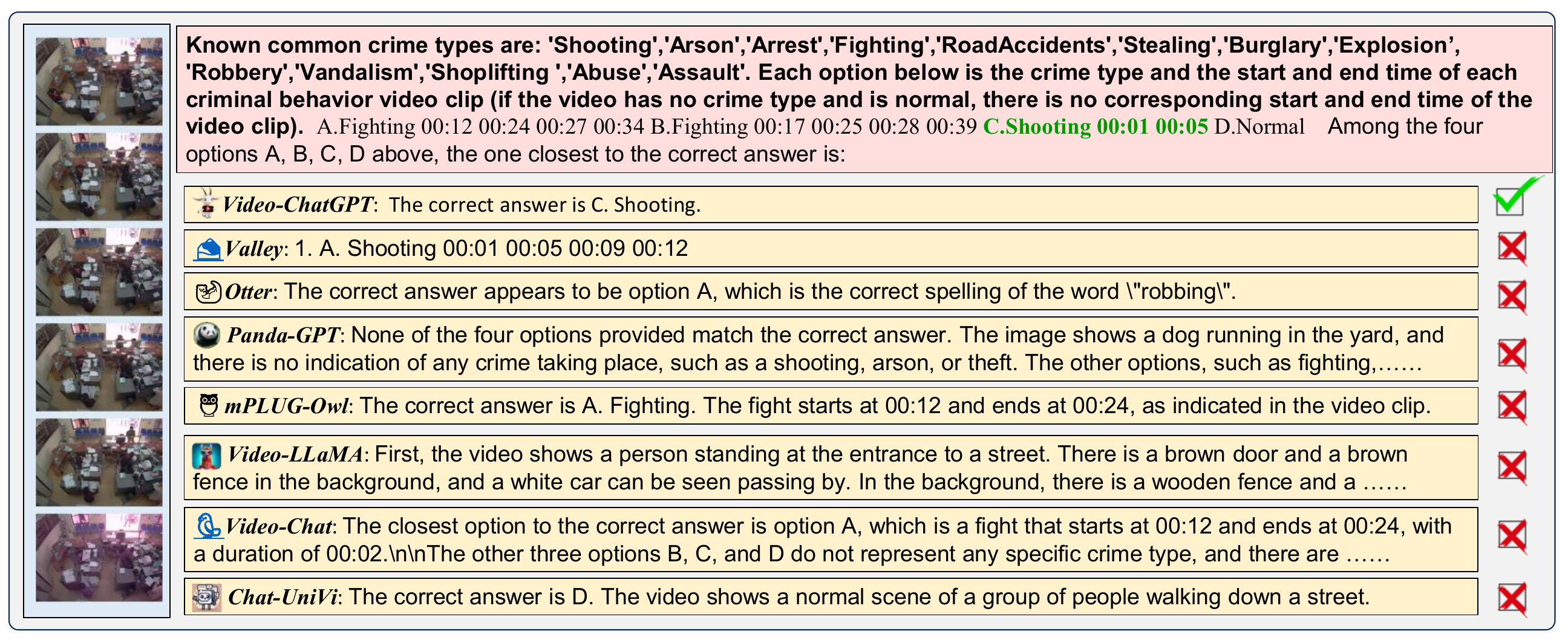}
  \caption{The illustrative sample of the generated responses from different Video-LLMs for UCF-Crime.}
  \label{fig:sup_ucf}
\end{figure*}

\vspace{0.1cm}
\noindent\textbf{UCF-Crime.} The results of the UCF-Crime is shown in  Fig.~\ref{fig:sup_ucf}. The poor performance illustrates the existing Video-LLMs lack the ability of temporal perception again.

\begin{figure*}
    \centering
    \includegraphics[width=2\columnwidth]{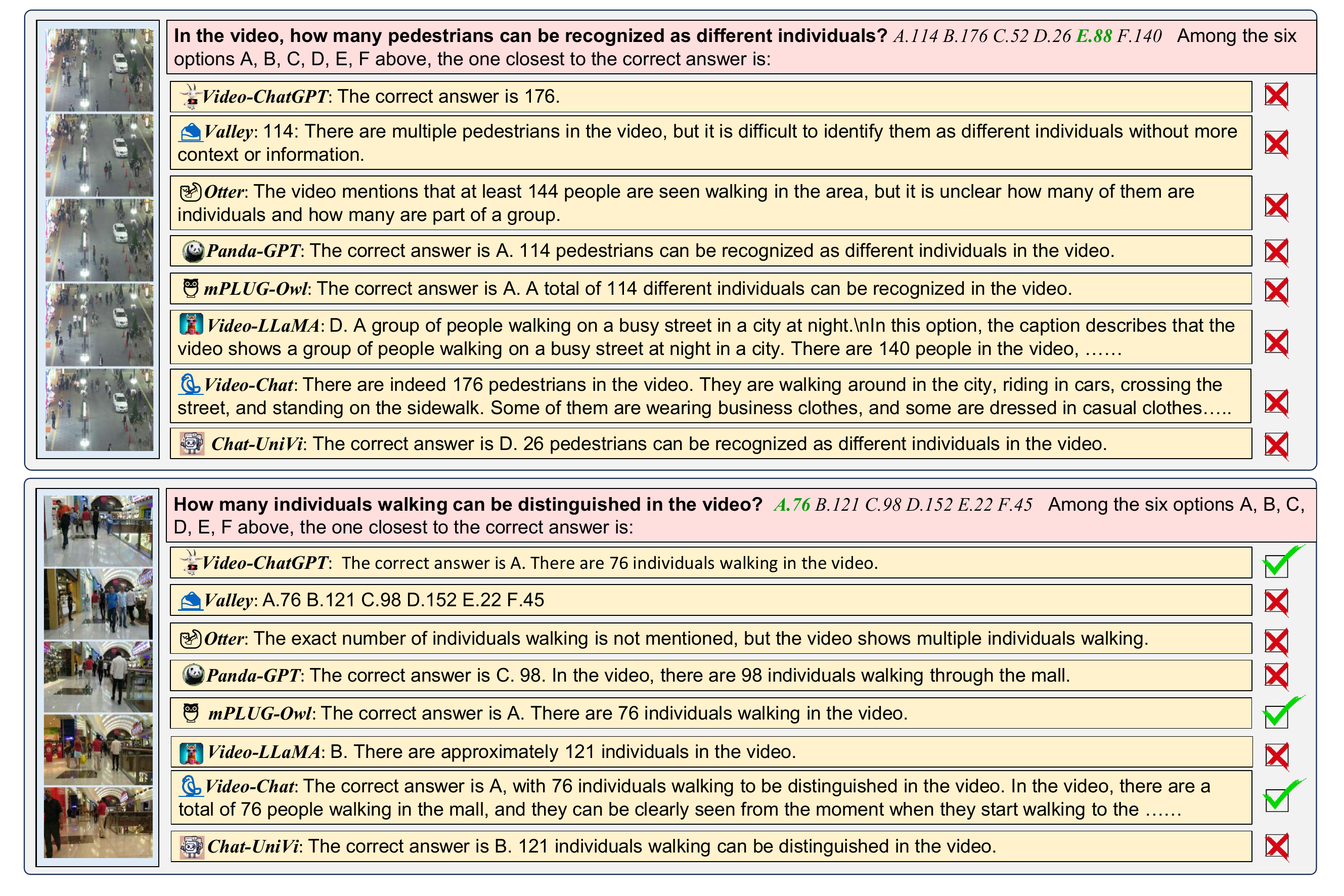}
    \caption{The illustrative sample of the generated responses from different Video-LLMs for MOT.}
    \label{fig:sup_mot}
\end{figure*}

\vspace{0.1cm}
\noindent\textbf{MOT.} The results of the MOT is shown in  Fig.~\ref{fig:sup_mot}. Existing Video-LLMs are proved to lack the ability to count accurately.

\subsection{Prior Knowledge-based Question-Answering}

\begin{figure*}
    \centering
    \includegraphics[width=2\columnwidth]{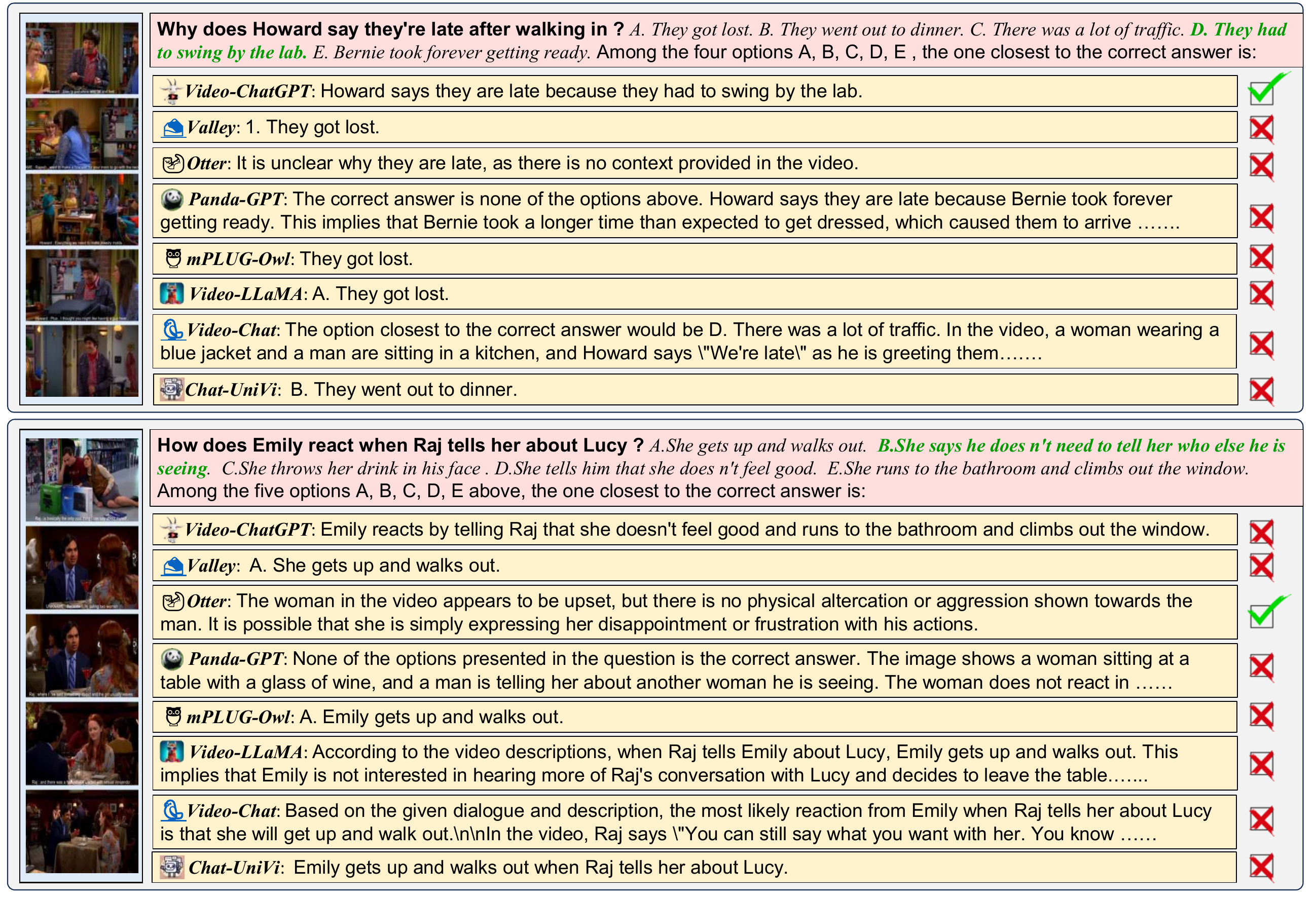}
    \caption{The illustrative sample of the generated responses from different Video-LLMs for TV-QA.}
    \label{fig:sup_tvqa}
\end{figure*}

\vspace{0.1cm}
\noindent\textbf{TV-QA.} The results of the TV-QA is shown in  Fig.~\ref{fig:sup_tvqa}, which demonstrate that existing Video-LLMs can hardly understand TV segments. This could be caused by the lack of prior knowledge and audio or subtitle understanding ability.

\begin{figure*}
  \centering
  \includegraphics[width=2\columnwidth]{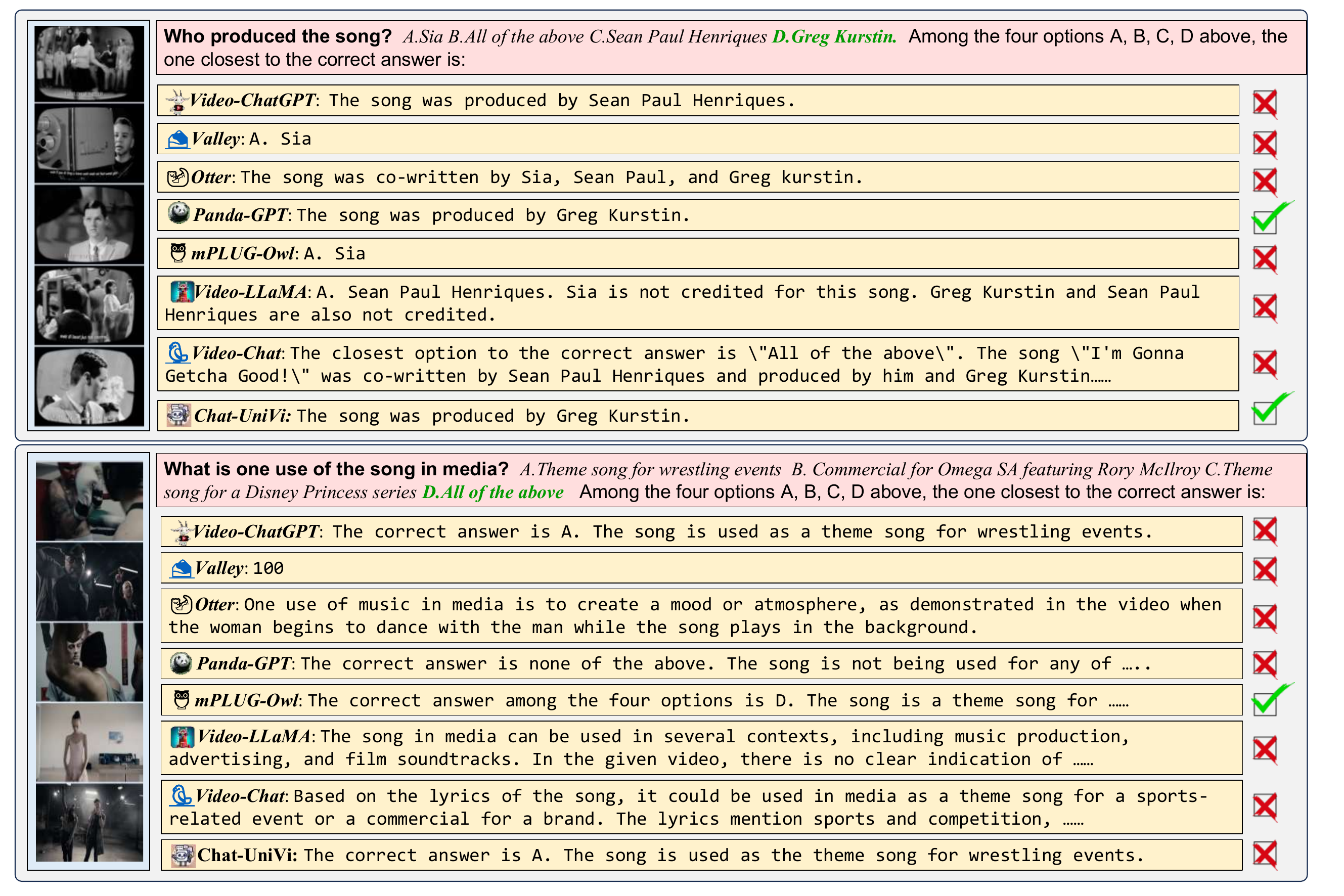}
  \caption{The illustrative sample of the generated responses from different Video-LLMs for MV-QA.}
  \label{fig:sup_mvqa}
\end{figure*}

\vspace{0.1cm}
\noindent\textbf{MV-QA.} The results of the MV-QA is shown in  Fig.~\ref{fig:sup_mvqa}. The poor performance may be also caused by the lack of prior knowledge and audio understanding ability.

\begin{figure*}
    \centering
    \includegraphics[width=2\columnwidth]{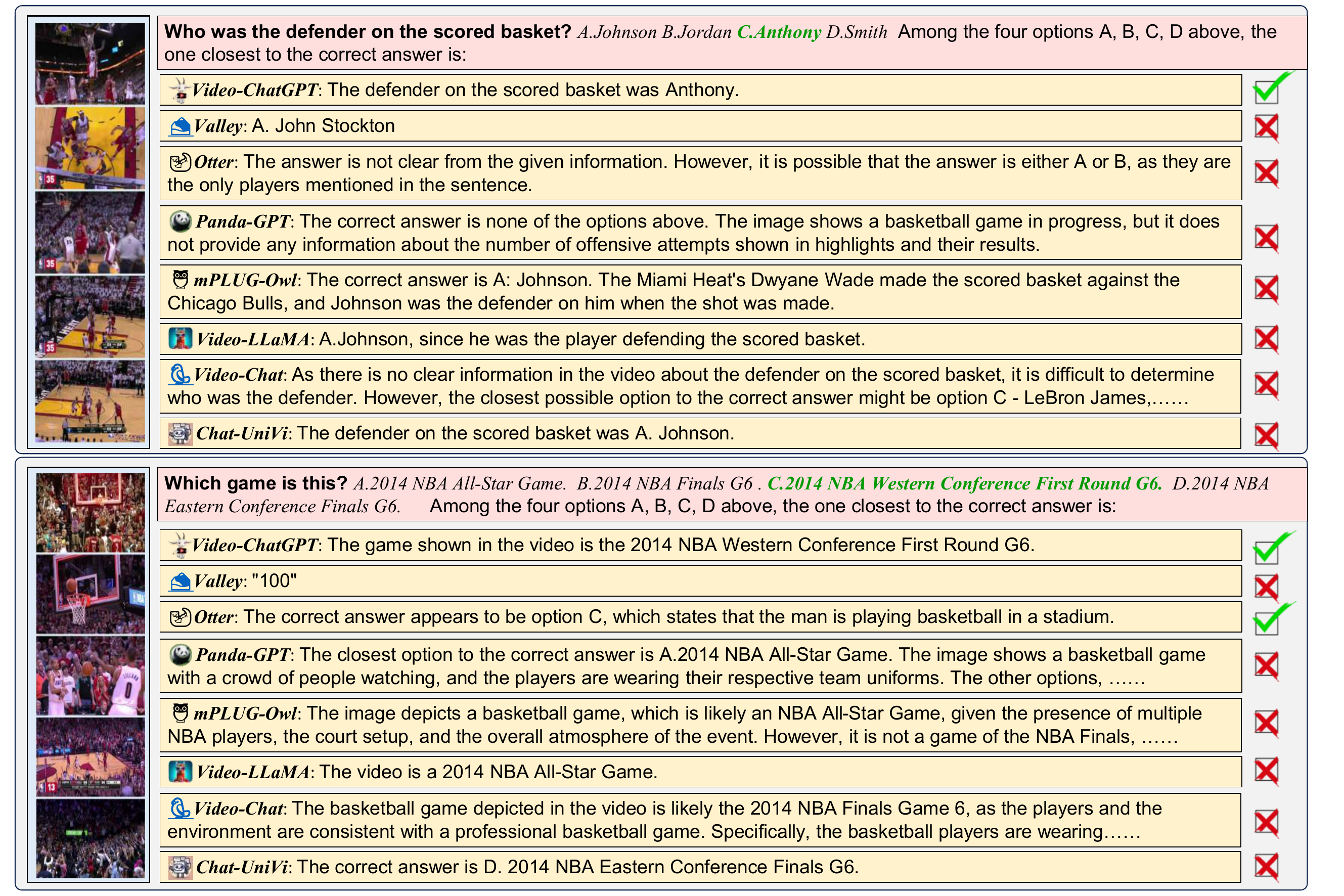}
    \caption{The illustrative sample of the generated responses from different Video-LLMs for NBA-QA.}
    \label{fig:sup_nba}
\end{figure*}

\vspace{0.1cm}
\noindent\textbf{NBA-QA.} The results of the NBA-QA is shown in  Fig.~\ref{fig:sup_nba}, which illustrates that without vision-language pre-training for specific domains, the Video-LLMs can not connect the knowledge stored in LLM with visual content and response to corresponding questions.

\subsection{Comprehension and Decision-Making}

\begin{figure*}
    \centering
    \includegraphics[width=2\columnwidth]{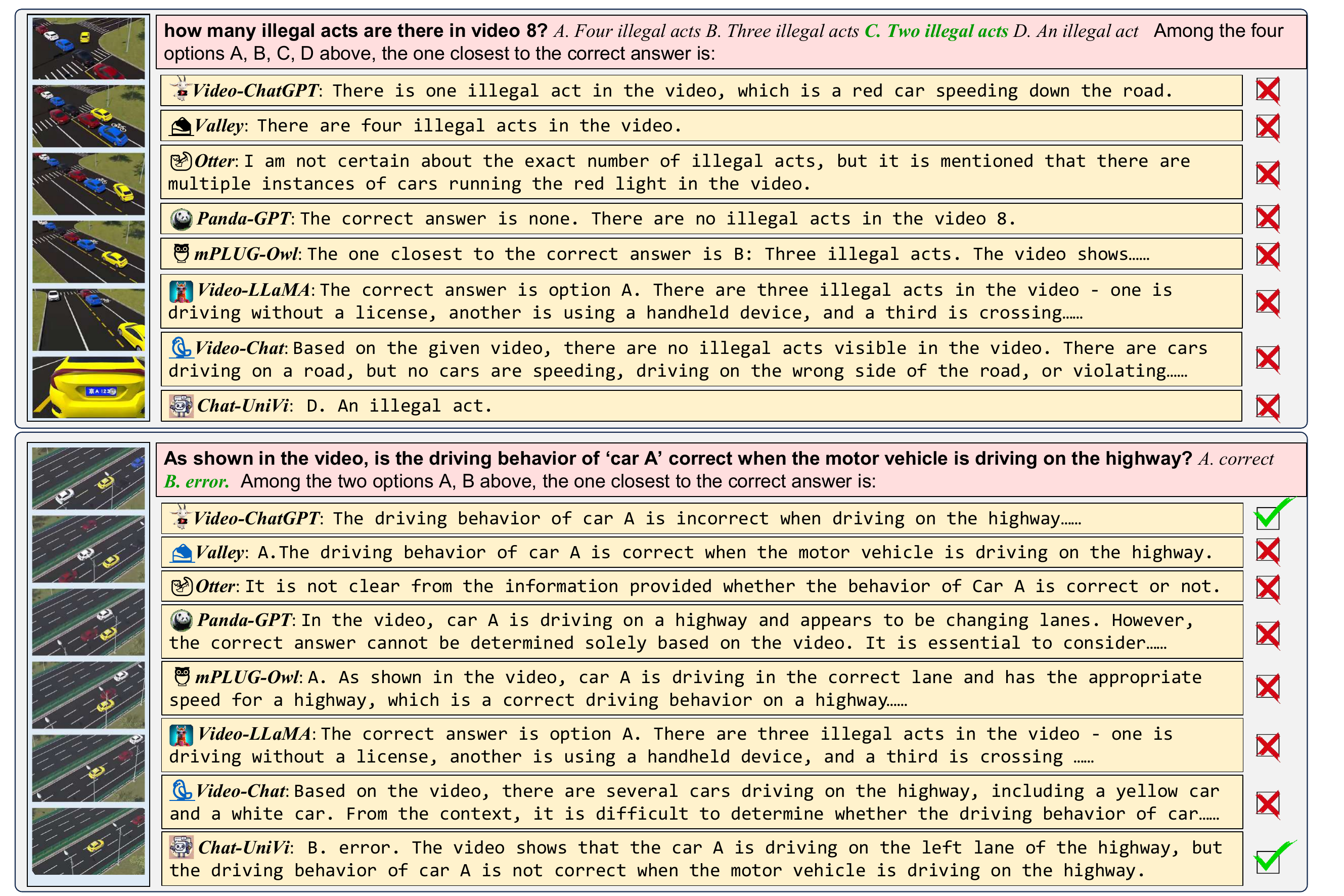}
    \caption{The illustrative sample of the generated responses from different Video-LLMs for Driver's License Examination.}
    \label{fig:sup_license}
\end{figure*}

\vspace{0.1cm}
\noindent\textbf{Driver's License Examination.} The results of the Driver's License Examination is shown in  Fig.~\ref{fig:sup_license}. The poor performance validates the tested Video-LLMs have limited scene understanding and decision-making ability.

\begin{figure*}
  \centering
  \includegraphics[width=2\columnwidth]{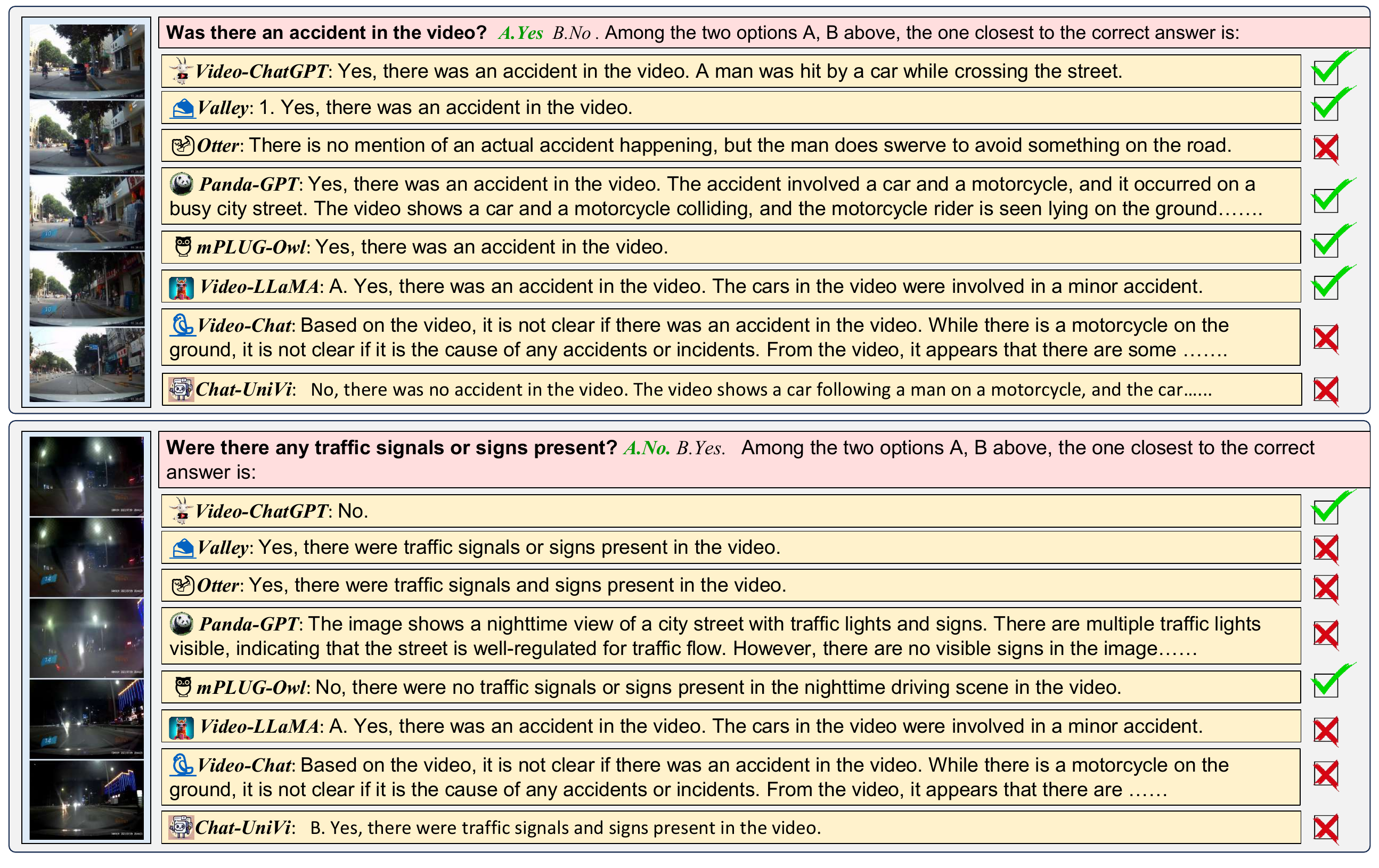}
  \caption{The illustrative sample of the generated responses from different Video-LLMs for Driving Decision-Making.}
  \label{fig:sup_decision}
\end{figure*}

\vspace{0.1cm}
\noindent\textbf{Driving Decision-Making.} The results of the Driving Decision-Making is shown in  Fig.~\ref{fig:sup_decision}, which demonstrates the tested Video-LLMs are difficult to understand the real driving environment.

\begin{figure*}
    \centering
    \includegraphics[width=2\columnwidth]{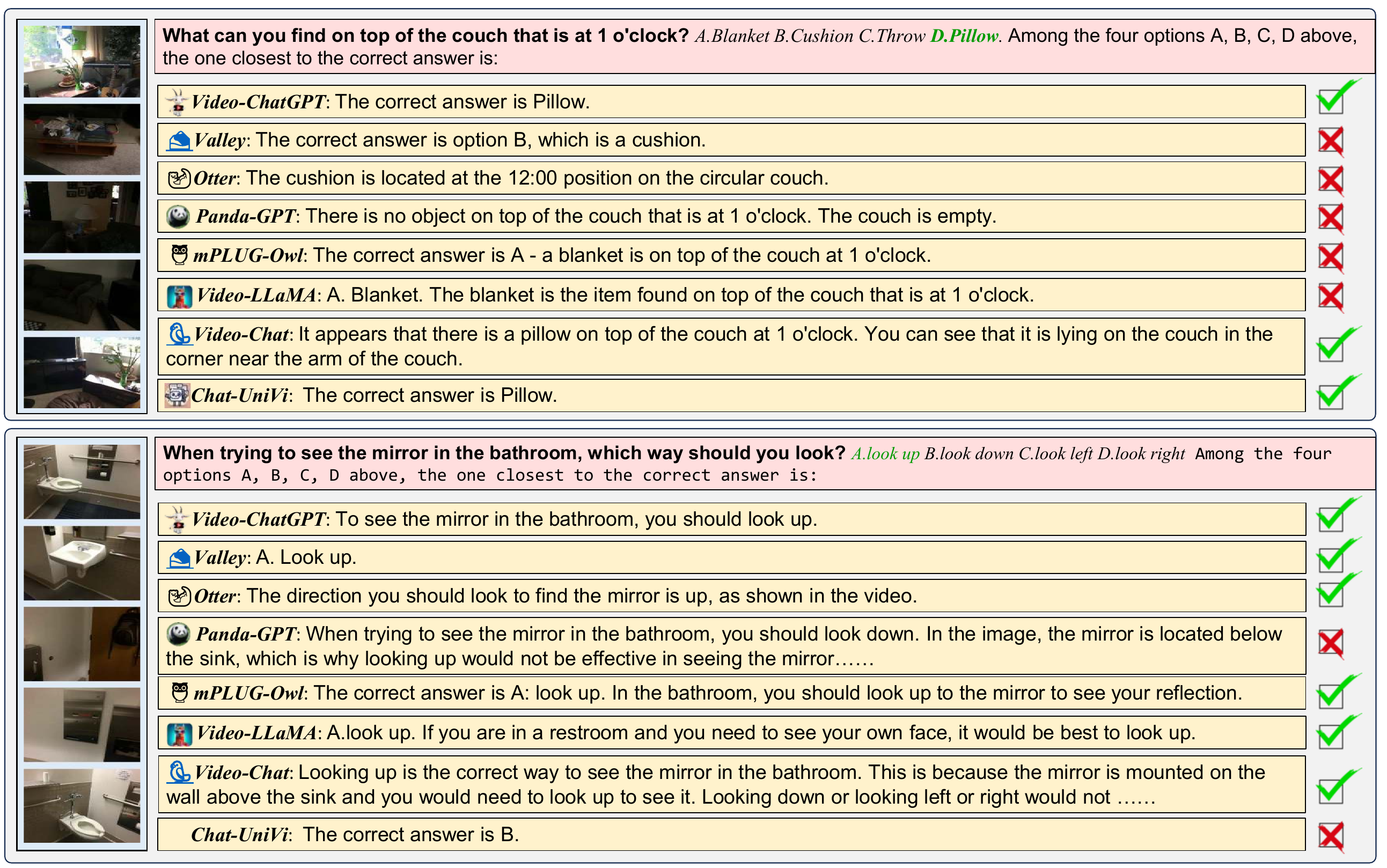}
    \caption{The illustrative sample of the generated responses from different Video-LLMs for SQA3D.}
    \label{fig:sup_sqa}
\end{figure*}

\vspace{0.1cm}
\noindent\textbf{SQA3D.} The results of the SQA3D is shown in  Fig.~\ref{fig:sup_sqa}. The results show that they can only understand the simple environment and cannot understand the complex spatial relationship.

\end{document}